\documentclass[acmsmall,screen, authoryear,table]{acmart}

\AtBeginDocument{%
  }
    
\usepackage{algorithm}
\usepackage{algorithmicx}
\usepackage{algpseudocode}
\usepackage{lineno}
\usepackage{amsmath}
\usepackage{xspace}
\newcommand{\nick}{\textit{IRepair}\xspace}
\newcommand{\nickkl}{\textit{IRepair + KL}\xspace}
\usepackage{graphicx}
\usepackage{multirow}
\usepackage{xcolor}
\usepackage{caption}
\usepackage[T1]{fontenc}
\usepackage{natbib}
\usepackage{bm}
\usepackage{subcaption}

\usepackage[normalem]{ulem}
\useunder{\uline}{\ul}{}
\setcopyright{cc}
\setcctype{by}
\acmDOI{10.1145/3715775}
\acmYear{2025}
\acmJournal{PACMSE}
\acmVolume{2}
\acmNumber{FSE}
\acmArticle{FSE056}
\acmMonth{7}
\received{2024-09-13}
\received[accepted]{2025-01-14}

\begin{document}

\title{\nick: An Intent-Aware Approach to Repair Data-Driven Errors in Large Language Models}

\author{Sayem Mohammad Imtiaz}
\orcid{0000-0002-0357-0098}
\affiliation{%
  \institution{Iowa State University}
  \city{Ames}
  \country{USA}
}
\email{sayem@iastate.edu}

\author{Astha Singh}
\orcid{0009-0000-3321-445X}
\affiliation{%
  \institution{Iowa State University}
  \city{Ames}
  \country{USA}
}
\email{asthas@iastate.edu}

\author{Fraol Batole}
\orcid{0000-0002-4586-1686}
\affiliation{%
  \institution{Tulane University}
  \city{New Orleans}
  \country{USA}
}
\email{fbatole@tulane.edu}

\author{Hridesh Rajan}
\orcid{0000-0002-9410-9562}
\affiliation{%
  \institution{Tulane University}
  \city{New Orleans}
  \country{USA}
}
\email{hrajan@tulane.edu}

\renewcommand{\shortauthors}{Sayem Mohammad Imtiaz, Astha Singh, Fraol Batole, Hridesh Rajan}
\begin{abstract}
Not a day goes by without hearing about the impressive feats of large language 
models (LLMs), 
and equally, not a day passes without hearing about their challenges.
LLMs are notoriously vulnerable to biases in their dataset, leading to issues 
such as toxicity, harmful responses, and factual inaccuracies. 
While domain-adaptive training has been employed to mitigate these issues, 
these techniques often address all model parameters indiscriminately
during the repair process, resulting in poor repair quality and reduced model versatility. In this paper, drawing inspiration from fault localization via program slicing, we introduce a novel dynamic slicing-based intent-aware LLM repair strategy, \nick. This approach selectively targets the most error-prone sections of the model for repair. Specifically, we propose dynamically slicing the model's most sensitive layers that require immediate attention, concentrating repair efforts on those areas. This method enables more effective repairs with potentially less impact on the model's overall performance by altering a smaller portion of the model. Furthermore, dynamic selection allows for a more nuanced and precise model repair compared to a fixed selection strategy. We evaluated our technique on three models from the GPT2 and GPT-Neo families, with parameters ranging from 800M to 1.6B, in a toxicity mitigation setup. Our results show that \nick repairs errors 43.6\% more effectively while causing 46\% less disruption to general performance compared to the closest baseline, \textit{direct preference optimization}. Our empirical analysis also reveals that errors are more concentrated in a smaller section of the model, with the top 20\% of layers exhibiting 773\% more error density than the remaining 80\%. This highlights the need for selective repair. Additionally, we demonstrate that a dynamic selection approach is essential for addressing errors dispersed throughout the model, ensuring a robust and efficient repair.

\end{abstract}

\ccsdesc[300]{Software and its engineering~Maintaining software}
\ccsdesc[100]{Computing methodologies~Natural language generation}
\ccsdesc[300]{Computing methodologies~Neural networks}

\keywords{SE4AI, Dynamic Program Slicing, Fault Localization, Program Repair, Large Language Model, Detoxification}

\maketitle

\section{Introduction}
\label{sec:intro}
The recent advancement in large language model (LLM) capabilities marks a transformative moment in natural language processing (NLP). Owing to the effectiveness of \textit{transformer} in scaling efficiently to the large corpus, LLMs now excel in tasks such as question-answering, text summarization, and code generation~\cite{zhao2023survey}. However, despite their impressive capabilities, LLMs are not without their shortcomings. Akin to traditional software, LLMs can exhibit bugs or generate unintended outputs, manifesting as toxicity, harmful responses, factual errors, or hallucinations~\cite{ji2023survey,korbak2023pretraining}. 
The root cause of these issues often lies in the training data itself~\cite{ji2023survey}. LLMs are typically trained on vast, unfiltered datasets, primarily sourced from the internet, using semi-supervised learning techniques~\cite{zhao2023survey}. It is impractical to validate such a vast corpus, eliminating biases, factual inconsistencies, and other issues~\cite{ji2023survey}. As a result, LLMs inadvertently inherit and propagate these issues in their outputs.

Existing approaches to mitigate such issues in LLMs primarily operate in three stages~\cite{wang2022exploring,korbak2023pretraining,pan2023automatically}: the pre-training stage~\cite{korbak2023pretraining,liu2024exposing}, alignment stages following pre-training, also known as domain-adaptive training (DAT) methods~\cite{wang2022exploring,gehman2020realtoxicityprompts,lee2024mechanistic,liu2023chain,rafailov2024direct}, and runtime methods~\cite{leong2023self,yang2022unified,xu2022leashing,niu2024parameter}.
Runtime methods, while impactful, often address symptoms rather than the root causes of model errors and can introduce computational overhead, limiting their applicability in low-latency scenarios~\cite{wang2022exploring,korbak2023pretraining,dathathri2019plug,huang2023survey}. Pre-training methods, such as Liu et al.'s attention-sharpening regularizer~\cite{liu2024exposing}, though effective, are costly due to the need for training from scratch, making them more suitable for new models~\cite{wang2022exploring,huang2023survey}. DAT methods, on the other hand, directly repair pre-trained models and provide an offline approach to error mitigation, making them particularly useful in real-time applications and situations requiring more invasive repair procedures~\cite{wang2022exploring,huang2023survey}.

DAT has two main paradigms: fine-tuning the model with curated data and preference optimization, such as reinforcement learning from human feedback (RLHF)~\cite{ouyang2022training}. However, both approaches update model parameters indiscriminately without considering their relevance to the problem at hand. This can decrease the effectiveness of the repair and increase the likelihood of negatively impacting the model's general performance by altering unrelated parameters. To address this, we introduce \nick, a dynamic slicing-based technique for selectively repairing only the intended part of the model.

Our motivation for targeted LLM repair is inspired by the successful application of the `fault localization followed by program repair' paradigm in traditional software engineering (SE). This approach has demonstrated effectiveness in producing optimal repairs while preserving the program's original structure as much as possible~\cite{mechtaev2015directfix,nguyen2013semfix,wen2018context}. For instance, Mechtaev \textit{et al.} employ partial MaxSAT constraint solving and component-based program synthesis to localize bugs and generate repairs, focusing on minimizing alterations to the program's structure~\cite{mechtaev2015directfix}. Inspired by these works, we adapt and evaluate this paradigm in the context of LLMs to address data-driven errors. Specifically, we propose a method that first localizes the source of errors within the model and then selectively repairs it. This approach aims to produce optimal repairs while preserving model performance by targeting only the relevant parts of the model and leaving unrelated sections unaffected.

To localize the source of errors within the model, we build upon the concept of relevant program slicing in software engineering~\cite{weiser1984program}. Relevant slicing identifies a subset of program statements that could impact a specified slicing criterion~\cite{gyimothy1999efficient, zhang2022remos}. Similarly, we apply these principles to identify and isolate the parts of the model that are most relevant to the errors being addressed.

Recently, slicing techniques have been adapted for deep learning models, offering advantages such as model protection and simplification~\cite{zhang2020dynamic} and vulnerability mitigation during transfer learning~\cite{zhang2022remos}. These approaches use a subset of data as the `slicing criteria' and analyze the model's activations to identify relevant parts of the model. However, these techniques rely on activation values to determine relevance, which is not directly applicable to the transformer architecture used in LLMs (as discussed in \S~\ref{sec:approach}). Additionally, these methods are more akin to static slicing, where the relevant sections of the model are selected after training. This results in a 'fixed selection,' which can be useful for various applications as demonstrated in previous works~\cite{zhang2020dynamic, zhang2022remos}. In contrast, we hypothesize that a dynamic selection technique, applied during the training process, will enable a more nuanced and precise repair of LLMs through domain-adaptive training with curated data.

Inspired by these works, \nick treats faulty data as a `slicing criterion' to identify error-prone sections of the model during each training pass. By analyzing the gradients of parameters with respect to the negative log-likelihood (NLL) of the faulty data, we pinpoint the components responsible for the unintended faulty responses. We then repair only the identified area while freezing the rest of the model, subject to a Kullback-Leibler (KL) divergence~\cite{csiszar1975divergence} constraint. This approach allows for focused repair efforts on the most critical sections, minimizing disruption to the existing knowledge stored in most model parameters. Moreover, \nick employs a dynamic slicing approach for repairing LLMs, enabling more nuanced and adaptive model repair compared to existing slicing methods that pre-select a fixed area.

To evaluate the effectiveness of our proposed technique, we conducted a case study focused on mitigating toxicity in LLMs. Given their pre-training on extensive corpora in a semi-supervised manner, LLMs are known to perpetuate biases and toxicity present in the data~\cite{xu2022leashing}. To assess the efficacy of our repair approach, we detoxified three models from the GPT-2 and GPT-Neo families, ranging from 800M to 1.6B parameters, using \nick. We compared our results against state-of-the-art baselines, employing the pairwise detoxification dataset developed by Lee \textit{et al.}~\cite{lee2024mechanistic}. Specifically, our baselines include representative techniques from different paradigms within domain-adaptive methods, such as Domain-Adaptive Pretraining (DAPT)~\cite{gururangan2020don,gehman2020realtoxicityprompts}, DAPT with a regularization term on the pre-training mixture to retain general performance during repair~\cite{liu2023chain}, and Direct Preference Optimization (DPO)~\cite{rafailov2024direct}, an RL-based preference optimization technique. Note that this study focuses on evaluating pre-trained language models (PLMs) that have not undergone further alignment. The term `LLM' used throughout the paper specifically refers to these PLMs. In this context, the general performance of the model refers to the quality of language generation from these models, measured in terms of perplexity metric.

We summarize the key contributions and findings of this paper as follows:

\begin{itemize} 
\item We propose using sensitivity as a measure of relevance for slicing transformer-based language models, addressing architectural challenges unique to these models.

\item Our framework not only facilitates targeted repair of LLMs but also adapts dynamically during training by slicing the model as needed. 
\item Unlike prior techniques, our method introduces a threshold-free slicing approach, eliminating the need for costly tuning, which can be expensive for large models such as LLMs.

\item Our analysis shows that the source of errors can be more pronounced in specific areas of the model than in others, and targeted interventions can deliver more efficient repairs compared to indiscriminate approaches.
\item Our approach, \nick, significantly outperforms state-of-the-art techniques, demonstrating greater efficiency in error elimination while preserving model generation quality. Specifically, \nick reduces toxicity by 43.6\% more than the closest baseline, \textit{DPO}, while causing 46\% less disruption to overall performance.

\item We also demonstrate that a dynamic selection approach is essential for addressing errors dispersed throughout the model, ensuring a robust and efficient repair.
\end{itemize}

\section{Background}
\label{sec:background}

LLMs have revolutionized NLP and SE tasks due to their ability to capture complex patterns and generate human-like text and code. In this section, we provide an overview of the GPT (Generative Pre-trained Transformer) architecture, which serves as a foundation for many modern LLMs~\cite{zhao2023survey}, including the models used in our study.

The GPT architecture, introduced by OpenAI \cite{radford2019language}, is based on the transformer model \cite{vaswani2017attention}. It consists of multiple layers of transformer blocks, which are the fundamental units of computation in the model. These blocks are also the primary focus of our slicing technique in \nick. Each block contains two main components:

\begin{itemize}
    \item \textbf{Multi-Head Attention}: This mechanism allows the model to focus on different parts of the input sequence simultaneously. Multi-head attention splits the input into multiple `heads,' each learning to attend to different input aspects. The attention function is defined as:

    \begin{equation}
        \text{Attention}(Q, K, V) = \text{softmax}\left(\frac{QK^T}{\sqrt{d_k}}\right)V
    \end{equation}
    
    \noindent where $Q$, $K$, and $V$ are query, key, and value matrices, respectively, obtained from linear projections of the input sequence. $d_k$ is the dimension of the key vectors. The attention output is then passed through a feed-forward network (FFN).

    \item \textbf{Feed-Forward Neural Network (FFN)}: This component processes the output of the attention mechanism. It is a neural network that operates on each position in the input sequence independently. The FFN typically consists of two linear transformations with a non-linear activation function in between:
    
    \begin{equation}
        \text{FFN}(x) = \max(0, xW_1 + b_1)W_2 + b_2
    \end{equation}
    
    \noindent where $W_1$, $W_2$, $b_1$, and $b_2$ are learnable parameters.
    
\end{itemize}

Each transformer block applies layer normalization and residual connections around these components.

\section{Approach}
\label{sec:approach}

\begin{figure}[]
	\centering
 \includegraphics[width=0.7\linewidth,trim=0cm 0cm 0cm 0cm]{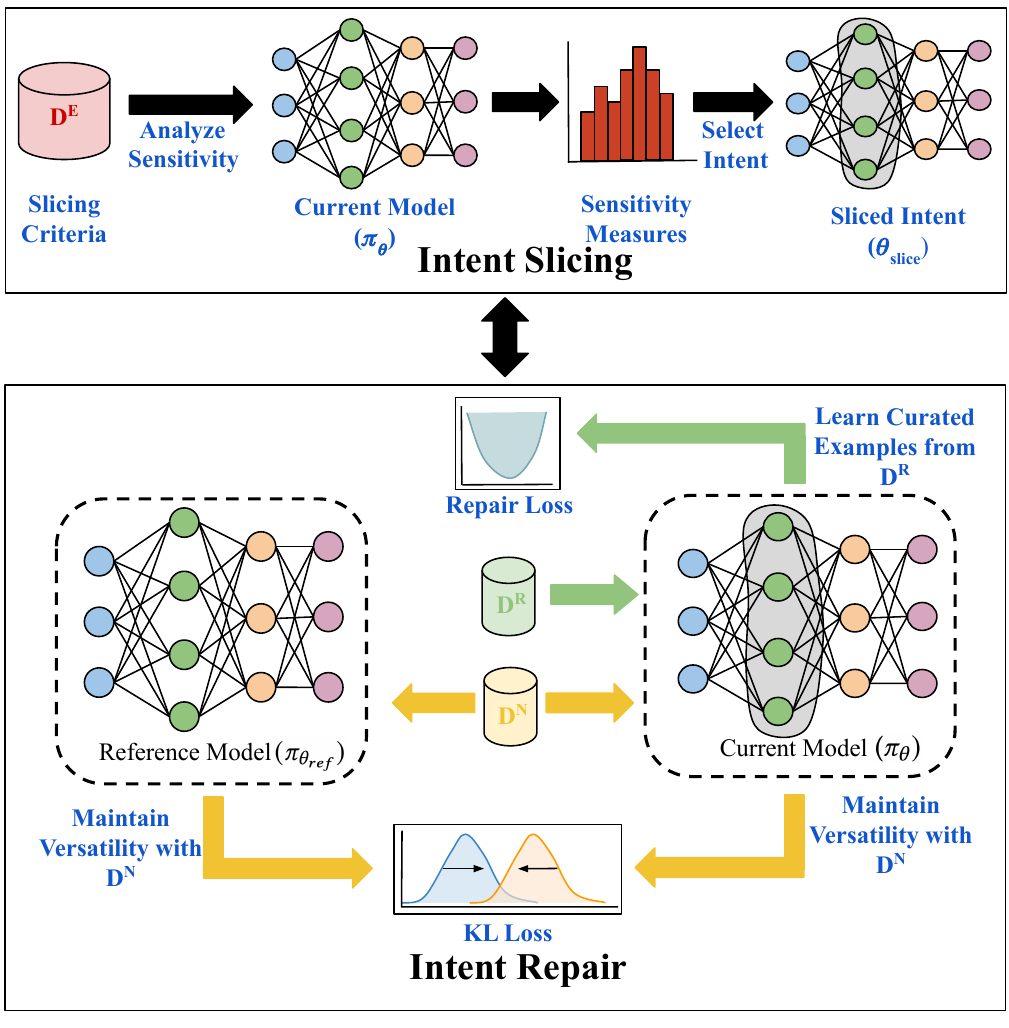}
	\vspace{-10pt}
	\caption{Overview of \nick.}
	\label{fig:overview}
\end{figure}

Figure~\ref{fig:overview} presents a high-level overview of the approach used in \nick, which aims to repair data-driven errors in large language models. The approach consists of two primary components: computing the relevant model slice and then repairing the identified slice selectively. In the first stage, we apply the concept of program slicing to language models to identify the slice that requires repair. In the second stage, we repair the identified slice selectively, focusing on the most error-prone sections of the model while minimizing any impact on its general performance. The following sections will explore the detailed steps involved in \nick.

\subsection{Problem Formulation}

Let \( \pi_\theta: X \rightarrow Y \) denote a pre-trained language model that maps input texts from \( X \) and a set of parameters, $\theta$, to corresponding output texts in \( Y \). Consider a bad demonstration dataset, denoted as \( D^E = (X^E, Y^E) \), where the response \( Y^E \) to a prompt \( X^E \) is undesirable. The goal is to ensure that the model \( \pi_\theta \) does not produce responses similar to \( Y^E \) when given prompts similar to \( X^E \). Additionally, consider a curated or refined dataset, denoted as \( D^R = (X^E, Y^R) \), which demonstrates the desirable response \( Y^R \) for those error-evoking prompts \( X^E \). In practice, these refined responses can be obtained through various methods, such as human annotation or conditioning a language model~\cite{solaiman2021process,lee2024mechanistic, wang2022exploring, pan2023automatically}. Once the refined data is acquired, the model \( \pi_\theta  \) is typically repaired via domain-adaptive training either by directly optimizing the model with the curated dataset~\cite{lee2024mechanistic,wang2022exploring,gehman2020realtoxicityprompts}, or the model is explicitly trained to prefer good demonstrations over bad ones via preference optimization~\cite{rafailov2024direct}.

However, repairing the errors in LLMs often involves a trade-off between overall model performance and repair quality~\cite{korbak2023pretraining,wang2022exploring}. Higher repair quality tends to come at the cost of reduced general performance~\cite{ding2022delta}. Our experiments also observed this trade-off across various techniques. Domain-adaptive repair methods are particularly susceptible to this issue because they update model parameters indiscriminately without considering their relevance to specific errors. This indiscriminate updating can lead to inefficient error repair, where the reduction in general performance may not justify the extent of error correction achieved.

To address these challenges, we propose focusing repairs on the sections of the model with the highest concentration of errors while leaving unrelated parameters unchanged. This approach aims to enhance repair efficiency, potentially achieving greater repair quality with less disruption to general performance compared to "intent-unaware" or indiscriminate techniques.

To that end, in this paper, we propose using examples from the bad demonstration dataset to identify and slice the most relevant sections of the model, referred to as intent, for selective repair. Specifically, we aim to pinpoint the most error-prone blocks within the transformer architecture for targeted intervention. Additionally, we hypothesize that errors may not be confined to a single block. To accommodate this, we introduce a dynamic slicing mechanism that allows for the selection of the most error-prone sections during the course of training. This approach enables a more nuanced and precise repair of errors throughout the model.

\subsection{Slicing Intent}

Drawing upon the concept of relevant program slicing~\cite{weiser1984program}, in this step, we aim to slice off the most error-prone sections of the model for a selective repair. Such a focused repair approach is critical for LLMs, as updating all parameters using limited repair data may lead to overfitting and knowledge degradation~\cite{ding2022delta}. By selectively slicing the model based on bad data, our aim is to address the root cause of errors in the model while preserving the model's overall knowledge.

\subsubsection{Challenges in Slicing LLM}
Existing slicing techniques for deep learning models are primarily designed for networks using \textit{ReLU} activations~\cite{zhang2022remos,zhang2020dynamic}. These techniques rely on activation status or their magnitudes to identify relevant parts of the model. However, transformer-based language models employ an attention mechanism with linear transformations, making these methods inapplicable.  In contrast to \textit{ReLU} activations, the magnitude of a linear transformation in \textit{attention} doesn't directly correspond to its importance due to subsequent matrix multiplications. As an illustration, consider the following simplified example of unmasked attention scores for a sequence with three tokens, $T_1, T_2$ and $T_3$ and an embedding dimension, $d_k=1$:

\[
Q=
  \begin{bmatrix}
    3 \\
    -3 \\
    2
  \end{bmatrix}
  \text{ and }
  K=
  \begin{bmatrix}
    -5 \\
    2 \\
    1
  \end{bmatrix}
\]

\begin{align*}
\text{Score} &= \sigma\left(\frac{QK^T}{\sqrt{d_k}}\right) = \sigma\left(
  \begin{bmatrix}
    -15 & 6 & 3\\
    15 & -6 & -3 \\
    -10 & 4 & 2
  \end{bmatrix}\right) \\
&=
   \begin{bmatrix}
    0 & 0.95 & 0.05\\
    1 & 0 & 0 \\
    0 & 0.88 & 0.12
  \end{bmatrix} \text{ (simplified)}
\end{align*}

Here, we observe that in predicting the next token for $T_2$, $T_1$ is the most influential. Thus, the multiplication of two negative values of \textit{Q} and \textit{K}, $-3 \times -5$, yields the highest score for $T_2$. Unlike \textit{ReLU}, where a node value less than zero might indicate irrelevance~\cite{zhang2022remos}, in this context, negative values do not reliably signify insignificance. Additionally, considering the magnitude of the activation level is inapplicable~\cite{zhang2020dynamic}, as the sign plays a crucial role in the score computation.

In addition, existing techniques require careful calibration of a threshold to select the slice~\cite{zhang2022remos,zhang2020dynamic}, which is a challenging task for large-scale models with billions of parameters, such as LLMs. Tuning for the optimal threshold can be very challenging and time-consuming for such models, highlighting the need for a threshold-free slicing approach for LLMs.

\subsubsection{Our Approach}
To address these challenges, we propose a \textit{gradient-based} approach for determining the relevance of model parameters to the slicing criterion, which does not rely on the activation of neurons. Specifically, we identify the relevant transformer block, referred to as \textit{intent}, of the model by assessing the sensitivity of the blocks to the slicing criterion. The approach works by treating a sample of bad data, ($D^E=(X^E, Y^E)$),  as criteria for slicing the intent that requires fixing. Figure~\ref{fig:overview} shows the overview of our proposed algorithm for slicing the LLM, which involves two major steps: assessing the relevance of parameters to the slicing criteria and intent or slice selection, which will be discussed below:

\begin{algorithm}
\caption{Slicing Intent}
\label{algo:compute_nll}
\begin{algorithmic}[1]
\Function{NLL}{$\pi_\theta$, $X$, $Y$}

  \State $logits \gets \pi_\theta(X)$ \label{algocn:2}\algorithmiccomment{Obtain last layer output}
  \State $logits \gets logits[:,-1,:]$ \label{algocn:3}\algorithmiccomment{Omit last token}
  \State $P \gets \sigma(logits)$ \label{algocn:4}\algorithmiccomment{Perform softmax operation}
 \State $mask \gets Y\neq \pi_{\theta}.padding\_id$ \label{algocn:1} 

  \State $n_1 \gets -Y \times \log(P) $ \label{algocn:5}\algorithmiccomment{NLL for tokens}
  
  \State $n_2 \gets \sum^{T}_{t}n_1(t) \times mask$ \label{algocn:6}\algorithmiccomment{NLL for sequence}
  \State $nll \gets \frac{n_2}{\sum^{T}_{t}mask(t)}$ \label{algocn:7}\algorithmiccomment{Mean NLL}

  \State \textbf{return} $nll$ \label{algocn:8}\algorithmiccomment{Return NLL}
\EndFunction

\Function{Sensitivity}{$\pi_\theta$, $X$, $Y$}
  \State $S \gets \{\}$ \label{algocs:2}
    \State $L \gets \text{NLL}(\pi_\theta, X, Y)$\label{algocs:3}\algorithmiccomment{Get Negative Log Likelihood}
    \For {every $block \in \pi_\theta$}\label{algocs:4}
        \State $S\{block\} \gets \sqrt{\sum_{i}^{\vartheta \in \theta_{block}} \left(\frac{\partial \mathcal{L}}{\partial w_i}\right)^2}$\label{algocs:5}\algorithmiccomment{L2-norm}

    \EndFor
  \State \textbf{return} $S$\label{algocs:6}
\EndFunction

\Function{Slice}{$\pi_\theta$,  $X$, $Y$}
  \State $S \gets \text{Sensitivity}(\pi_\theta, X, Y)$\label{algoid:1}\algorithmiccomment{Measure sensitivity}
  \State $B \gets \operatorname*{arg\,max}_{block \in \pi_\theta} S\{block\}$ \label{algoid:2}\algorithmiccomment{Get most error-prone block}
\State $\theta_\text{slice} \gets \{\vartheta | \vartheta \in \theta_B \land \theta_B \in \theta\}$ \label{algoid:3}\algorithmiccomment{Get slice}
  \State  \textbf{return} $\theta_\text{slice}$\label{algoid:4}
\EndFunction
\end{algorithmic}
\end{algorithm}

\paragraph{Computing sensitivity to slicing criteria}
As previously discussed, activation-based approaches are not applicable in the context of the \textit{transformer}. Instead, we identify the \textit{intent} by assessing the sensitivity of blocks to \textit{slicing criteria}. To that end, we propose leveraging \textit{negative log-likelihood (NLL)} of the model response to assess the relevance of blocks to the slicing criteria. Specifically, we first measure the impact of all parameters on the model's response by calculating the first-order gradient of the \textit{NLL} of the generated response. Then, we compute the sensitivity of the block by taking the \textit{L2-norm} of the gradients of all parameters within the block. Without loss of generality, the sensitivity of a block to a slicing criterion, $x$, can be represented as:

\[
\label{eq:sen1}
S\{\text{block}\} \approx \left\| \nabla_{\theta_{\text{block}}} \left( -\sum_{t=1}^{T} \log p_{\theta}(x_t \mid x_{1:t-1}) \right) \right\|_2
\]

Here, \( \mathbf{\theta}_{\text{block}} \) represents the parameters within a specific block of the transformer, \( p_\theta(x_t \mid x_{1:t-1}) \) represents the probability of the token \( x_t \) given the prior tokens \( x_{1:t-1} \), \( T \) denotes the total number of tokens in \( x \), and \( \mathbf{\theta} \) refers to the overall parameter space. The notation \( \| \cdot \|_2 \) denotes the L2 norm, which is applied to the gradient of the NLL with respect to the block's parameters.

The \textit{NLL} reflects the model's confidence in generating the target response. A lower NLL score indicates higher confidence in accurately predicting the target response. Taking the gradients of parameters with respect to the \textit{NLL} provides a measure of their sensitivity to the inputs or criteria provided. If a small increase or perturbation in a parameter leads to a noticeable impact on the model output, that parameter is likely important or relevant to the criteria~\cite{kirkpatrick2017overcoming}. The greater the magnitude or norm of the gradient for a parameter, the more relevant it is to the criteria.

The \textit{Sensitivity} method (in Algorithm~\ref{algo:compute_nll}) provides our approach for computing sensitivity to slicing criteria. The method takes an instance of the model ($\pi_{\theta}$) and slicing criteria as $X$ and $Y$. It first calculates the \textit{negative log-likelihood} by invoking the \textit{NLL} method in line~\ref{algocs:3}.

To achieve this, the \textit{NLL} method first obtains the model logits (the output of the last layer) by forward passing the input through the model (Line~\ref{algocn:2}). The last generated token is then discarded, as the corresponding token in the ground truth response does not exist (Line~\ref{algocn:3}). Using the \textit{softmax} activation function, the probability distribution for all output tokens in the vocabulary is calculated (Line~\ref{algocn:4}). To compute the \textit{NLL}, a loss mask is obtained where the special padding tokens are skipped to eliminate their impact on the calculation (Line~\ref{algocn:1}). The \textit{NLL} for each token in the sequence is calculated individually by multiplying the probability distribution of ground truth response ($Y$) with the log-likelihood of the output ($\log(P)$) (Line~\ref{algocn:5}). Next, the \textit{NLL} for the entire sequence is calculated by summing the individual NLLs for every token in the sequence, with padding tokens eliminated by multiplying by the \textit{mask} (Line~\ref{algocn:6}). Finally, in Line~\ref{algocn:7}, the mean \textit{NLL} is computed by dividing by the count of non-padding tokens in the sequence. The final averaged \textit{NLL} approximates the model's confidence in generating the target responses, $Y$, for the given inputs, $X$, and returned from the method as the final outcome.

In the \textit{Sensitivity} method, after obtaining the \textit{NLL} for the given criteria, the magnitude of the gradients for each transformer block is calculated by taking the \textit{L2 norms} of all parameters within the block with respect to the \textit{NLL} (Lines~\ref{algocs:4}–\ref{algocs:5}). Specifically, the first-order gradients for all the model parameters are computed with respect to the \textit{NLL} (i.e., $\nabla_\theta (L)$). For each parameter within a transformer block ($\vartheta \in \theta_{\text{block}} \land \theta_{\text{block}} \subset \theta$), the gradients are squared, and the square root of their summation is taken to yield the overall magnitude or sensitivity of the block ($S$) (Line~\ref{algocs:4}). This measure indicates the relevance of the block to the given slicing criteria and is used in the \textit{Slice} method to identify and slice the most error-prone block.

\paragraph{Selecting Intent}

The final method, \textit{Slice}, in Algorithm~\ref{algo:compute_nll}, slices the most error-prone block of the model for the provided criteria by leveraging the other two methods. The method takes as input an instance of the model, denoted by \( \pi_{\theta} \), where $\theta$ represents the parameter space of the model, and a set of slicing criteria, \( X \) and \( Y \). It first computes the sensitivity of every block in the model by invoking the \textit{Sensitivity} method (Line~\ref{algoid:1}). Then, in Line~\ref{algoid:2}, the block with the highest sensitivity—deemed most relevant to the provided criteria—is selected. When the provided criteria correspond to a sample from poor demonstration data \( D^E \), this block represents where the most error is concentrated. This step effectively eliminates the need for thresholding to identify the slice. Next, in Line~\ref{algoid:3}, the parameters within the selected block are sliced off and returned by the method.

\begin{algorithm}
\caption{Repairing Intent}
\label{algo:repair}
\begin{algorithmic}[1]
\Function{Repair}{$\pi_\theta$, $\pi_{\theta_{ref}}$, $D^E$, $D^R$, $D^N$, $\alpha$}
  \Repeat
    \State $X^R, Y^R \gets \text{batch}(D^R)$\label{algorp:1}\algorithmiccomment{Get a good batch}
    \State $X^E, Y^E \gets \text{batch}(D^E)$\label{algorp:2}\algorithmiccomment{Get corresponding bad batch as slicing criteria}
    \State $X^N, Y^N \gets \text{batch}(D^N)$\label{algorp:3}\algorithmiccomment{Get a normal batch}

    \State $\theta_\text{slice} \gets \text{Slice}(\pi_\theta,X^E, Y^E)$\algorithmiccomment{Sliced parameters}\label{algorp:4}

    \State $L_1 \gets \text{NLL}(X^R, Y^R, \theta_{\text{slice}})$\algorithmiccomment{Repair loss}\label{algorp:5}
    \State $L_2 \gets \text{KL}(X^N, Y^N, \theta_{\text{slice}}, \theta_{\text{ref}})$\algorithmiccomment{KL loss}\label{algorp:6}
    \State $L \gets \alpha \cdot L_1 + L_2$\label{algorp:7}\algorithmiccomment{Total loss}
    \State $G \gets \nabla(L, \theta_\text{slice})$\algorithmiccomment{Gradients w.r.t slice}\label{algorp:8}
    \State $update(\theta_{\text{slice}}, G)$\algorithmiccomment{Update parameters of $M$}\label{algorp:9}
  \Until{convergence}\label{algorp:10}
  \State \textbf{return} $\pi_{\theta}$\algorithmiccomment{Return repaired model}\label{algorp:11}
\EndFunction
\end{algorithmic}

\end{algorithm}

\subsection{Repairing Intent}

In this step, the identified slice or intent, $\theta_{\text{slice}}$, which is primarily responsible for undesirable generation, is addressed through two optimization objectives, as shown in Figure~\ref{fig:overview}. Specifically, our loss function includes an NLL term as repair loss and a KL term to preserve the model's normal utility, as shown in Equation~\ref{eq:rpob}.

\begin{equation}
\label{eq:rpob}
\begin{aligned}
Loss = & \alpha \cdot \text{NLL}\left( p_{\theta_{\text{slice}}}(\cdot \mid X^R) \right) \\
& + KL\left( p_{\theta_{\text{slice}}}(\cdot \mid X^N) \parallel p_{\theta_{\text{ref}}}(\cdot \mid X^N) \right)
\end{aligned}
\end{equation}

Here, \( \alpha \) represents the strength of the repair or NLL loss, \( \theta_{\text{slice}} \) denotes the set of sliced parameters from the model \( \pi_\theta \), \( \theta_{\text{ref}} \) represents the parameter space of the reference model \( \pi_{\theta_{\text{ref}}} \), and \( p_{\theta_{\text{slice}}} \) denotes the probability distribution of \( \pi_\theta \) conditioned on the sliced parameters. The key components of the repair process are briefly described below:

\subsubsection{Repair Loss}

We use the negative log-likelihood (\textit{NLL}) as the repair loss for our technique. This loss aims to maximize the log-likelihood of the curated responses (\(Y^R\)) for fault-evoking prompts (\(X^E\)). \textit{NLL} is a commonly employed loss function for repairing models via continued pre-training or supervised fine-tuning~\cite{wang2022exploring, gehman2020realtoxicityprompts, lee2024mechanistic,geva2022transformer}. However, unlike existing techniques, we only optimize the sliced parameters (\(\theta_{\text{slice}}\)) of the patient model, \(\pi_\theta\). This selective approach allows for more focused and aggressive updates of the error-prone parameters, potentially leading to more effective repairs. Additionally, updating a smaller portion of the total parameters reduces general performance degradation, as most of the model retains its original parameters. The relative importance of this term is regulated by the $\alpha$ coefficient. 

\subsubsection{KL Loss}
\textit{KL loss} is used to preserve the general performance of the model during the repair process. Specifically, this term aims to minimize the divergence between a reference distribution (the output of the reference model, $\pi_{\theta_{\text{ref}}}$) and the target distribution (output of $\pi_{\theta}$ on the pre-training corpus \(D^N\)). This term essentially encourages the model to maintain similar generation capabilities to the reference model on unrelated aspects.

\subsubsection{Dynamic Slicing}
Finally, we employ a dynamic slicing mechanism that selects the most error-prone block of the model during the course of training for an adaptive repair. This design decision is motivated by three key factors:
\paragraph{Error concentration} First, our threshold-free slicing technique selects only the most relevant or error-prone block of the model based on the criteria. However, a single block may not be solely responsible for undesirable responses to certain prompts. Other parts of the model might also significantly contribute to erroneous outputs, as we empirically confirm in our analysis. We find that errors can span multiple blocks, necessitating the repair of more than one block (details in \S~\ref{sec:rq3}). In such cases, repairing only a predetermined fixed block may not be sufficient.

\paragraph{Error movement} Second, a fixed selection strategy, like those used in existing works~\cite{zhang2020dynamic, zhang2022remos}, may fail to adapt to the effects of training dynamics on the model. While an area of the model might appear most responsible for undesirable responses before repair, it may not remain the most error-prone block throughout the course of training. Once training adequately addresses the initially selected area, another unselected area may appear more problematic, deserving more repair effort at that point. A dynamic slicing technique that accounts for the impact of training dynamics can more effectively address such shifts in error concentration.

\paragraph{Local error correction}

As demonstrated in our algorithm for repairing intent (Algorithm~\ref{algo:repair}), we use corresponding bad examples for each batch of good examples (Line~\ref{algorp:1}) from the bad demonstration dataset, \( D^E \), as criteria to slice the most relevant block of the model (Line~\ref{algorp:2}). This approach ensures that repair efforts focus on the block that most amplifies errors for the current batch of data. In contrast to pre-selection strategies, which often use all or a sample of data to determine which part to slice~\cite{zhang2020dynamic, zhang2022remos}, our method enables a more nuanced repair by allowing for localized error correction.

\subsubsection{Algorithm Overview}

The \textit{Repair} method in Algorithm~\ref{algo:repair} outlines the procedure for repairing the intent using our proposed optimization objectives. The method takes as input the affected or to-be-repaired model \( \pi_\theta \), a reference model \( \pi_{\theta_{\text{ref}}} \), which is the same model as initial \( \pi_\theta \) and is used to maintain similar performance on unrelated aspects post-correction, and references to bad examples (\(D^E\)), good examples (\(D^R\)), and normal examples (\(D^N\)). Additionally, a hyper-parameter \( \alpha \) is provided, which is used as a measure of the strength of the repair loss.

The repair process begins by sampling a batch of good examples in Line~\ref{algorp:1} during each training iteration. It then constructs a batch of corresponding bad examples to use as slicing criteria for the current iteration (Line~\ref{algorp:2}). Additionally, a random batch from the normal examples is obtained to compute a KL term (Line~\ref{algorp:3}). Next, the \textit{Slice} method is invoked to extract the most error-inducing block of the current model, \( \pi_\theta \) (Line~\ref{algorp:4}). The \textit{NLL} loss for the sliced parameters, \( \theta_{\text{slice}} \), is computed using the good batch in Line~\ref{algorp:5}. Similarly, a KL term is calculated for both the currently repaired model and the reference model using the batch of normal examples (Line~\ref{algorp:6}). In Line~\ref{algorp:7}, the combined loss is obtained, with the \textit{NLL} term regulated by a user-defined coefficient (\(\alpha\)). After computing the loss, gradients with respect to the sliced parameters are computed in Line~\ref{algorp:8} and updated in Line~\ref{algorp:9}. The repair process continues until convergence or early stopping is triggered and the repaired model is returned.

\section{Evaluation}
\label{sec:evaluation}

In this section, we introduce our evaluation setup, outline our research questions, and discuss the experimental results in detail. As previously mentioned, we evaluate our technique within a model detoxification framework, where our goal is to repair toxic models using the principles outlined in \nick. To this end, we examine our framework across three research questions:

\begin{itemize} 
    \item \textbf{RQ1:} \textit{How effectively can \nick repair or detoxify the model?} This research question evaluates \nick's effectiveness in eliminating toxicity from models and compares it against several state-of-the-art baseline techniques.

    \item \textbf{RQ2:} \textit{What is the computational overhead of \nick?} This research question measures the computational overhead of \nick by assessing total floating point operations (FLOPS), peak memory usage, and convergence duration and compares these metrics against baseline techniques.
    
    \item \textbf{RQ3:} \textit{Does the dynamic selection employed by \nick offer any advantage?} In this research, we conduct ablation studies and empirical analyses of error concentration in the model to assess the impact and necessity of selective and dynamic repair.
\end{itemize}

\subsection{Experimental Setup}

\subsubsection{Model}
\label{sec:model}

We evaluate \nick across three models from the GPT family, namely GPT-2 Large (812M parameters), GPT-2 XL (1.61B parameters), and GPT-Neo (1.3B parameters). The GPT-2 models, developed by \textit{OpenAI}, were trained on 8 million web pages from the \textit{WebText} dataset~\cite{radford2019language}. The GPT-Neo model, developed by \textit{EleutherAI}, was trained on the \textit{PILE} dataset~\cite{gptneo}. 
We load these pre-trained models from the official Hugging Face repositories of OpenAI and EleutherAI~\cite{gptneohf,gpthf}.

\subsubsection{Dataset}

To evaluate \nick, we used the detoxification dataset developed by Lee \textit{et al.}~\cite{lee2024mechanistic}, consisting of 24,576 toxic and non-toxic pairs generated by GPT-2 models~\cite{radford2019language} from prompts sampled from the WikiText-2 train split~\cite{merity2016pointer}. The dataset is balanced with an equal number of toxic and non-toxic examples for detoxification. In our experimental setup, we treat the toxic continuations as a bad demonstration dataset, $D^E$, and the non-toxic continuations as a good dataset, $D^R$. The evaluation data includes a test set and a small development sample, which consists of 50 challenge prompts from REALTOXICITYPROMPTS~\cite{gehman2020realtoxicityprompts} and a subset of the WikiText-2 test split, totaling around 32.7K tokens. This development set is used for tuning hyperparameters through repeated runs with various combinations.

Additionally, for each model, we construct an individual normal dataset, \(D^N\), a curated pre-training subset that preserves diversity while minimizing targeted error symptoms. We leverage the notion of `unconditional generation' to construct this dataset~\cite{wang2022exploring}. Specifically, starting with the special start-of-sequence token (GPT models use `<|endoftext|>` as the start-of-sequence token~\cite{wang2022exploring}), we generate approximately 15,000 texts for each model using different random seeds. Following prior work, we employ nucleus sampling with a temperature of 1 and \(p = 0.9\) during generation~\cite{wang2022exploring}. The unconditionally generated text corpus is considered a good representative of the model's training corpus~\cite{wang2022exploring}. Then, we score each generation using perspective API and remove the ones with toxicity scores higher than 0.5~\cite{gehman2020realtoxicityprompts}, removing approximately 1.2\% of the initial dataset.
By minimizing the KL divergence in these examples, we aim to preserve the model's ability to generate random text similarly to its pre-repair state, thereby reducing the impact on its general performance.

\subsubsection{Baseline}
\label{sec:baseline}

We compare the performance of two variants of \nick: the standard \nick, which does not enforce a KL constraint, and \nickkl, which does, against several representative state-of-the-art baselines within domain-adaptive training, as introduced below:

\paragraph{Domain-Adaptive Pretraining (DAPT)} DAPT is a framework introduced by Gururangan \textit{et al.}~\cite{gururangan2020don}, which involves continuing the pretraining of a model on domain-specific texts. Gehman \textit{et al.}~\cite{gehman2020realtoxicityprompts} applied the DAPT framework to further train GPT-2 models on nontoxic texts to detoxify them. In our setup, we evaluate the effectiveness of DAPT in detoxifying models and compare it with \nick.

\paragraph{Direct Preference Optimization (DPO)} DPO is a cutting-edge algorithm designed to replace RLHF (Reinforcement Learning from Human Feedback~\cite{ouyang2022training}) due to its complex and unstable training process. It directly steers the model towards desirable generations over undesirable ones~\cite{rafailov2024direct}. The DPO algorithm has been shown to effectively eliminate toxicity from models, as demonstrated by Lee \textit{et al.}~\cite{lee2024mechanistic}. We also compare our method against DPO.

\paragraph{Domain-Adaptive Pretraining with KL Constraint (DAPT+KL)}
We also compare \nick against a variant of the DAPT method that includes a KL constraint to preserve the general model performance~\cite{liu2023chain}. While DAPT alone may cause the model to deviate when training on a domain-specific corpus, adding a KL term helps evaluate the repair quality of this approach by ensuring that the model maintains its general performance while focusing on domain-specific adjustments.

Additionally, in RQ3, we compare \nick against two of its variants to assess the effectiveness of the components employed by \nick, as described below:

\paragraph{\nick (Min)} In this variant of \nick, during each training iteration, instead of selecting the block with the highest error concentration, the block with the least concentration is chosen. Comparing \nick against this variant allows us to assess the impact of selective repair.

\paragraph{\nick (Fixed)} In this variant of \nick, a fixed slice of the model is pre-selected for repair. Specifically, using a substantial random sample of bad data ($D^E$) consisting of 2000 examples, average sensitivities for all parameters are computed. Then, using the same technique described in Algorithm~\ref{algo:compute_nll}, block sensitivity is computed, and the block with the highest sensitivity is selected for repair. Comparing \nick against this variant allows us to assess the impact of dynamic selection.

\subsubsection{Metric}

Following the prior works~\cite{gehman2020realtoxicityprompts,lee2024mechanistic,geva2022transformer}, we use the following two metrics to evaluate the toxicity and general performance of the model after detoxification.

\paragraph{Toxicity}Gehman \textit{et al.} developed a dataset called \textit{REALTOXICITYPROMPTS} to evaluate toxicity in LLMs~\cite{gehman2020realtoxicityprompts}. This dataset consists of sentence-level prompts that are provided to LLMs to generate continuations, which are likely to elicit toxic responses from the models. They also created a \textit{challenge} subset of this dataset, which includes 1,199 prompts that consistently caused all models in their experiments to generate toxic responses~\cite{gehman2020realtoxicityprompts}. This \textit{challenge} subset has been used in previous studies to evaluate the effectiveness of detoxification methods~\cite{lee2024mechanistic,geva2022transformer}. Similarly, we leverage this subset to assess the detoxification quality of our repaired models. Additionally, following prior works~\cite{geva2022transformer,lee2024mechanistic,gehman2020realtoxicityprompts}, we use the widely adopted toxicity detection tool, \textit{PERSPECTIVE API}~\cite{papi}, to assign a toxicity score to each generation. The score ranges from 0 to 1, with higher scores indicating more toxic responses. The toxicity score for the test data is calculated as the average toxicity score across the entire test set, following prior works~\cite{lee2024mechanistic}.

\paragraph{PPL} \textit{Perplexity} (PPL) is a widely used metric to evaluate the generation quality of language models. It has been employed to assess degradation in model generation after the repair process~\cite{gehman2020realtoxicityprompts, lee2024mechanistic, geva2022transformer, wang2022exploring}. \textit{PPL} measures how uncertain or "perplexed" the model is in predicting the next word. Higher PPL indicates that the model is worse at predicting the next word, meaning its generation quality is lower. When evaluated on a test corpus, it reflects how well the model's generated text aligns with the test data. Following prior works on GPT models~\cite{geva2022transformer, lee2024mechanistic}, we compute the \textit{PPL} of the model before and after repair using two benchmarks. \textit{PPL (WikiText2)} is computed on the test split of \textit{WikiText-2}~\cite{merity2016pointer}, which contains 4,358 rows and approximately 241K words. \textit{PPL (Lambada)} is evaluated on the LAMBADA benchmark~\cite{paperno2016lambada}, which assesses the text-understanding capabilities of language models. To calculate PPL, we split the test corpus into 1024-token segments, compute the NLL for target tokens in each segment, and weigh the NLL values by the number of target tokens. The perplexity is then computed as the exponentiation of the mean NLL per token.

Additionally, we evaluate the computational overhead of the techniques using the following three metrics, as described in the literature~\cite{kaplan2020scaling,shoeybi2019megatron,lee2019energy}:

\paragraph{TFLOPs} \textit{TFLOPs} (Tera Floating-Point Operations per Second) represents the total number of floating-point operations performed in the trillions during the course of training or inference~\cite{kaplan2020scaling}. This metric is commonly used to gauge the computational demand of a method. To compute the \textit{TFLOPs}, we utilize an open-source tool~\cite{casson2023transformerflops} that applies the formula derived by Kaplan \textit{et al.} for GPT models~\cite{kaplan2020scaling}.

\paragraph{Peak Memory Usage} Memory consumption is a critical factor when developing techniques for large language models (LLMs)\cite{shoeybi2019megatron}. To evaluate this, we measure the peak memory usage of all techniques during training. We employ the method used by Lee \textit{et al.}~\cite{lee2019energy}, where an independent process queries the GPU using the \textit{nvidia-smi} command at 1-second intervals to record the highest memory usage observed.

\paragraph{GPU Time} We report the total GPU time required for training each technique until convergence. This measurement is obtained using \textit{PyTorch's CUDA} API~\cite{torchcuda}, which tracks the time spent on the GPU throughout the training process.

\paragraph{Total Iteration} Additionally, we report the total iteration needed until the model converges or early stopping is triggered.


\subsubsection{Training Details}

As discussed in \S~\ref{sec:model}, we used pre-trained models from the official \textit{Hugging Face} repositories of \textit{OpenAI} and \textit{EleutherAI}~\cite{gpthf, gptneohf}. We leveraged Python's deep learning library \textit{PyTorch}~\cite{paszke2017automatic} for further training these models across all techniques. All hyperparameters in our study were fine-tuned using a small development dataset produced by Lee \textit{et al.}~\cite{lee2024mechanistic}. For \textit{DPO}, \textit{DAPT}, and \textit{DAPT + KL}, we used the implementation provided by Lee \textit{et al.}~\cite{lee2024mechanistic} as a reference. 

Similarly, we fine-tuned the hyperparameters for all techniques using the development set. Specifically, the final tuned learning rates for standard \nick, \nickkl, \textit{DPO}, \textit{DAPT}, and \textit{DAPT+KL} are $2e^{-5}$, $5e^{-5}$, $1e^{-6}$, $1e^{-6}$, and $5e^{-6}$, respectively. Through trial and error, we found that a higher learning rate tends to achieve better repair quality at the expense of general performance and vice versa. Since \nick only modifies a small portion of the model, it can accommodate a larger learning rate with less adverse impact on general performance compared to other indiscriminate techniques. Similarly, \textit{DAPT+KL} allows a slightly higher learning rate than \textit{DPO} and \textit{DAPT} as it explicitly aims to maintain general performance during repair. We also set the value of $\alpha$ to 0.5 for both \nick and \textit{DAPT+KL} after tuning.

For training models using all techniques, we used the memory-efficient \textit{RMSProp} optimizer with 150 warmup steps and a linear learning rate scheduler. A batch size of four was used for the techniques, with a validation split of eight batches, each with a batch size of eight. Models were trained with a validation loss patience of 30 iterations. All models were trained on an \textit{NVIDIA A100} GPU with 40GB of memory. We conducted all the training using the same random seed to ensure reproducibility and enable a fairer comparison.

\begin{table}[]
\centering
\footnotesize
\caption{Comparative Overview of \nick's Performance. (Toxicity scores are scaled from 0 to 100. The best performance is highlighted in bold, and the second-best is underlined for each model.}
\begin{tabular}{|c|c|c|c|c|c|c|c|}
\hline
\textbf{Model} &
  \textbf{Metric} &
  \textbf{Vanilla} &
  \textbf{DAPT} &
  \textbf{DAPT+KL} &
  \textbf{DPO} &
  \textbf{\nick} &
  \textbf{\nickkl} \\ \hline
                               & Toxicity   & 41.53 & 11.99 & 13.93          & 14.97 & {\ul 7.74}     & \textbf{5.11}  \\ \cline{2-8} 
& PPL (WikiText2) & 19.44 & 26.95 & 22.96          & 24.10 & \textbf{20.93} & {\ul 22.35}    \\ \cline{2-8}
\multirow{-3}{*}{GPT2 812M}    & PPL (Lambada) & 30.40 & 43.59 & 34.74          & 38.79 & {\ul 32.51} & {\textbf{30.93}}    \\ \hline

                               & Toxicity   & 48.22 & 39.74 & {\ul 5.21}     & 22.71 & 10.67          & \textbf{4.68}  \\ \cline{2-8} 
& PPL (WikiText2) & 17.40 & 22.96 & 20.13          & 21.02 & {\ul 18.16}    & \textbf{18.15} \\ \cline{2-8} 
\multirow{-3}{*}{GPT2 1.61B}   & PPL (Lambada) & 28.10 & 38.92 & 34.70          & 36.33 &  \textbf{29.19}    & {\ul 31.83} \\ \hline

                               & Toxicity   & 40.89 & 31.57 & 37.93          & 12.26 & {\ul 5.69}     & \textbf{4.94}  \\ \cline{2-8} 
 & PPL (WikiText2) & 14.56 & 17.00 & \textbf{16.17} & 16.70 & 16.56          & {\ul 16.52}     \\ \cline{2-8} 
\multirow{-3}{*}{GPT Neo 1.3B}   & PPL (Lambada) & 24.24 & 27.74 & 27.34          & \textbf{26.37} & 27.16    & {\ul 26.86} \\ \hline

\rowcolor[HTML]{EFEFEF} 
\cellcolor[HTML]{EFEFEF}       & Toxicity   & 43.6$\pm$2.3 & 27.8$\pm$8.2 & 19$\pm$9.8          & 16.7$\pm$3.1 & {\ul 8}$\pm$\underline{1.4}    & \textbf{4.9}$\pm$\textbf{0.1}  \\ \cline{2-8} 
\rowcolor[HTML]{EFEFEF} 

 &
  PPL (WikiText2)  &
  17.1$\pm$1.4 &
  22.3$\pm$2.9 &
  19.8$\pm$2.0 &
  20.6$\pm$2.1 &
  \textbf{18.6}$\pm$\textbf{1.3} &
  {\ul 19}$\pm$\underline{1.7}  \\ \cline{2-8} 
\rowcolor[HTML]{EFEFEF} 

\multirow{-3}{*}{\cellcolor[HTML]{EFEFEF}\textbf{Overall}} &
  PPL (Lambada)  &
  27.6$\pm$1.8 &
  36.8$\pm$4.7 &
  32.3$\pm$2.5 &
  33.8$\pm$3.8 &
  \textbf{29.6}$\pm$\textbf{1.6} &
  {\ul 29.9}$\pm$\underline{1.5} \\ \hline
  
\end{tabular}
\label{tab:rq1}
\end{table}
\subsection{RQ1: How Effectively Can \nick Repair the Model?}
\label{sec:rq1}

In this research question, we evaluate the repair effectiveness of \nick and compare it against several baselines. Table~\ref{tab:rq1} provides a comparative overview of \nick's performance across all models. The results show that both variants of \nick consistently outperform all other techniques on every model tested, achieving a higher repair score with better general performance stability. Specifically, standard \nick achieves an average 81.6\% reduction in toxicity, with an 8.3\% and 7.4\% increase in PPL on the \textit{WikiText2} and \textit{Lambada} benchmarks, respectively, across all models. The \nickkl variant reduces toxicity by 88.7\% with an 11\% and 8.3\% increase in PPL in \textit{WikiText2} and \textit{Lambada}. 

In contrast, \textit{DPO}, \textit{DAPT+KL}, and \textit{DAPT} reduce toxicity by 61.8\%, 56.3\%, and 36.2\%, while increasing \textit{PPL (WikiText2)} by 20.3\% (\textit{PPL (Lambada)}: 22.7\%), 15.3\% (\textit{PPL (Lambada)}: 17\%), and 30.2\% (\textit{PPL (Lambada)}: 33.3\%), respectively. Thus, both \nick variants clearly outperform all baseline techniques in both metrics. For example, compared to \textit{DPO}, standard \nick and \nickkl are 32\% and 43.6\% more effective in reducing toxicity while incurring 59.2\% and 46\% less increase in \textit{PPL (WikiText2)}. Similarly, against \textit{DAPT+KL}, standard \nick and \nickkl are 44.8\% and 57.5\% more effective in reducing toxicity while showing 45.8\% and 28.2\% less increase in \textit{PPL (WikiText2)}. Additionally, we find that all techniques, including \nick, significantly outperform \textit{DAPT}. A similar effect is observed in the \textit{PPL (Lambada)} benchmark as well.

Understandably, the ability of \textit{DAPT} to repair the model is limited by its tendency to lose general performance more rapidly. Overall, its PPL increases by 30.2\% (33.3\%) compared to increases of 15.3\% (17\%), 20.3\% (22.7\%), 8.3\% (7.4\%), and 11\% (8.3\%) for \textit{DAPT+KL}, \textit{DPO}, \nick,and \nickkl, respectively, in \textit{WikiText2} and \textit{Lambada}. This shows that adding a KL term to the \textit{DAPT} loss for self-generated random data helps better preserve unrelated model knowledge. Furthermore, the results demonstrate that compared to \textit{DAPT+KL}, which operates on all parameters indiscriminately, \nick's selective approach is more effective at controlling performance degradation with 45.8\% (56.4\%) and 28.2\% (51.2\%) less \textit{PPL (WikiText2)} (\textit{PPL (Lambada)}) increase than \textit{DAPT+KL}, for instance, despite using a higher learning rate. Although in the GPT-Neo model, PPL is slightly better for \textit{DAPT+KL} and \textit{DPO} on \textit{WikiText2} and \textit{Lambada}, \nick methods achieve better trade-offs between PPL and repair quality. For instance, the toxicity-to-PPL ratio for \textit{DAPT+KL} is 2.34 and 1.39 on \textit{WikiText2} and \textit{Lambada}, respectively, while for \textit{DPO}, it is 0.73 and 0.46. In contrast, standard \nick achieves 0.34 and 0.21, while \nickkl achieves 0.3 and 0.18—both lower than their counterparts—suggesting that \nick achieves a better trade-off between PPL and toxicity, even though a particular learning rate (LR) may result in higher PPL.

The results also demonstrate that KL-enabled techniques achieve better repair or toxicity scores, as they allow for more aggressive model repair at higher LR. For instance, \textit{DAPT+KL} reduces the toxicity score by 55\% compared to its non-KL counterpart, \textit{DAPT}. Similarly, \nickkl achieves a 9\% greater reduction in toxicity and exhibits 93\% lower standard error, indicating greater stability compared to standard \nick. Even with KL-enabled \textit{DAPT}, both \nick variants significantly outperform it in toxicity reduction by 44.8\% and 57.5\%, while incurring proportionally less disruption to general performance. \nick's ability to support higher LRs is a crucial factor in its effectiveness. However, this added efficiency is also largely driven by \nick's focused repair approach, as demonstrated in \S~\ref{sec:rq3}. Despite having a lower LR, the other approaches result in high PPL due to altering a large fraction of unrelated parameters. While a small LR implies minor individual changes, their cumulative effect can lead to a significant shift, causing high PPL. However, \nick focuses only on the areas with the highest error concentration, leaving most parameters untouched during each training pass. As a result, the model tends to converge 
on the error validation dataset before significantly affecting unrelated parameters, leading to highly precise adjustments and limiting large parameter shifts overall.

\begin{table}[]
\footnotesize
\centering
\caption{The Computational Overhead of the \nick compared with Baseline Techniques.}
\begin{tabular}{|c|c|c|c|c|c|c|}
\hline
\textbf{Model} &
  \textbf{Metric} &
  \textbf{DAPT} &
  \textbf{DAPT+KL} &
  \textbf{DPO} &
  \textbf{IRepair} &
  \textbf{IRepair + KL} \\ \hline
                               & TFLOPs                  & \textbf{12.32}          & 24.82 & {\ul 18.47}          & 45.97       & 63.19  \\ \cline{2-7} 
                               & TFLOPs/Token            & \textbf{4.92}           & 8.19  & 9.83           & {\ul 6.89}        & 10.17  \\ \cline{2-7} 
                               & GPU Time (sec)          & 1,994          & {\ul 1,411} & \textbf{733}           & 2,032        & 1,933  \\ \cline{2-7} 
                               & Peak Memory Usage (MiB) & \textbf{13175}          & 16547 & 18677          & {\ul 13990}       & 17071  \\ \cline{2-7} 
\multirow{-5}{*}{GPT2 812M}    & Total Iteration         & 6200           & {\ul 3750}  & \textbf{2325}           & 8250        & 5125   \\ \hline
                               & TFLOPs                  & {\ul 46.63}          & 65.34 & \textbf{35.24}          & 51.34       & 121.53 \\ \cline{2-7} 
                               & TFLOPs/Token            & \textbf{9.80}           & 16.33 & 19.60          & {\ul 13.52}       & 20.05  \\ \cline{2-7} 
                               & GPU Time (sec)          & 7,398          & 3,438 & \textbf{1,304}          & {\ul 2,010}        & 2,769  \\ \cline{2-7} 
                               & Peak Memory Usage (MiB) & {\ul 27664}          & 32411 & 35659          & \textbf{25042}       & 29333  \\ \cline{2-7} 
\multirow{-5}{*}{GPT2 1.61B}   & Total Iteration         & 11775          & 4950  & \textbf{2225}           & {\ul 4700}        & 5000   \\ \hline
                               & TFLOPs                  & \textbf{10.26}          & 19.39 & 27.37          & {\ul 19.03}       & 64.41  \\ \cline{2-7} 
                               & TFLOPs/Token            & \textbf{8.47}           & 14.12 & 16.94          & {\ul 11.92}       & 17.57  \\ \cline{2-7} 
                               & GPU Time (sec)          & 1,416          & {\ul 747}   & 1,013          & \textbf{545}         & 1,193  \\ \cline{2-7} 
                               & Peak Memory Usage       & {\ul 22842}          & 26233 & 25853          & \textbf{20356}       & 23179  \\ \cline{2-7} 
\multirow{-5}{*}{GPT Neo 1.3B} & Total Iteration         & 3000           & \textbf{1700}  & 2000           & {\ul 1975}        & 3025   \\ \hline

\rowcolor[HTML]{EFEFEF} 
\cellcolor[HTML]{EFEFEF}       & TFLOPs                  & \textbf{1.3} & 3.3 & {\ul 2.3}    & 3       & 5  \\ \cline{2-7} 
\rowcolor[HTML]{EFEFEF} 
\cellcolor[HTML]{EFEFEF}       & TFLOPs/Token            & \textbf{1}  & 3 & 4         & {\ul 2} & 5  \\ \cline{2-7} 
\rowcolor[HTML]{EFEFEF} 
\cellcolor[HTML]{EFEFEF}       & GPU Time           & 4.7          & {\ul 2.7} & \textbf{1.7} & {\ul 2.7}  & 3.3  \\ \cline{2-7} 
\rowcolor[HTML]{EFEFEF} 
\cellcolor[HTML]{EFEFEF} &
  Peak Memory Usage (MiB) &
  {\ul 1.7} &
  4 &
  4.7 &
  \textbf{1.3} &
  3.3 \\ \cline{2-7} 
\rowcolor[HTML]{EFEFEF} 
\multirow{-5}{*}{\cellcolor[HTML]{EFEFEF}\textbf{Avg. Ranking}} &
  Total Iteration &
  4.3 &
  {\ul 2} &
  \textbf{1.7} &
  3 &
  4 \\ \hline

\end{tabular}
\label{tab:rq2}
\end{table}

\subsection{RQ2: What Is the Computational Overhead of \nick?}
\label{sec:rq2}

In this research question, we evaluate the computational overhead of \nick and compare it against baseline techniques. Table~\ref{tab:rq2} presents the overhead of various techniques across four metrics. Among the two \nick variants, the overhead of standard \nick is more amenable to the other three baseline techniques across all metrics. For example, overall, it ranks first in memory consumption and second in GPU time despite incurring higher TFLOPs and requiring more iterations to converge.

The additional compute units (TFLOPs) consumed by \nick variants are due to the extra forward pass and a higher number of iterations required for convergence. \nickkl involves four forward passes: one for a toxic batch of data to assess the sensitivity ($D^E$), one for a non-toxic batch ($D^R$), and for normal data, one pass through the model under repair ($\pi_\theta$) and another through the reference model ($\pi_{\theta_{\text{ref}}}$). In contrast, standard \nick does not compute the KL term, thereby eliminating two forward passes for normal data, which results in significantly lower TFLOPs overall—53\% less than \nickkl. \textit{DPO} also requires four forward passes; however, it converges in fewer iterations, leading to lower total TFLOPs.

On a per-token basis, the TFLOPs required for \nick are comparable to other baselines (with standard \nick ranking second and \nickkl requiring the most, though comparable to \textit{DPO}). However, due to the higher number of iterations needed for convergence to address a smaller part of the model, total TFLOPs are higher. Despite this, \nick’s GPU time remains proportionally lower, and it trains faster or comparably to some baseline techniques, such as \textit{DAPT} and \textit{DAPT+KL}. This could be attributed to better GPU utilization in \nick variants, which process more TFLOPs per iteration, while \textit{DAPT} takes longer to converge, spreading out its TFLOPs and leading to lower average GPU utilization.

In terms of memory consumption, we find that standard \nick uses the least memory, while \nickkl ranks third. This is because the backward pass in \nick is more constrained than in other techniques. First, to calculate sensitivity, it only computes gradients for the transformer blocks, excluding the \textit{embedding} and final output layers. After slicing the layer, it zeroes out the gradients, freeing memory. In the second backward pass, it only computes gradients for the required smaller slice, resulting in lower peak memory consumption compared to other techniques. \nickkl requires slightly more memory than \textit{DAPT} due to storing extra logits for normal data from two forward passes.

Overall, while \nick incurs higher TFLOPs due to longer iterations, it remains memory-efficient and trains reasonably faster by fully utilizing available GPU power. In exchange for additional compute units, \nick offers better repair efficiency than the other techniques. Between \nick and \nickkl, although the latter is more computationally intensive, it provides greater stability in model repair, as observed in \S~\ref{sec:rq1}.

\begin{table}[]
\small
\centering
\caption{Impact of Dynamic Slicing on Repair Efficacy}
\begin{tabular}{|c|c|ccc|ccc|}
\hline
 &
   &
  \multicolumn{3}{c|}{\textbf{IRepair}} &
  \multicolumn{3}{c|}{\textbf{IRepair + KL}} \\ \cline{3-8} 
\multirow{-2}{*}{\textbf{Model}} &
  \multirow{-2}{*}{\textbf{Metric}} &
  \multicolumn{1}{c|}{\textbf{Fixed}} &
  \multicolumn{1}{c|}{\textbf{Min}} &
  \textbf{Original} &
  \multicolumn{1}{c|}{\textbf{Fixed}} &
  \multicolumn{1}{c|}{\textbf{Min}} &
  \textbf{Original} \\ \hline
 &
  Toxicity &
  \multicolumn{1}{c|}{45.53} &
  \multicolumn{1}{c|}{36.84} &
  7.74 &
  \multicolumn{1}{c|}{39.04} &
  \multicolumn{1}{c|}{42.08} &
  5.11 \\ \cline{2-8} 
\multirow{-2}{*}{GPT2 812M} &
  PPL (WikiText2) &
  \multicolumn{1}{c|}{20.78} &
  \multicolumn{1}{c|}{21.59} &
  20.93 &
  \multicolumn{1}{c|}{22.88} &
  \multicolumn{1}{c|}{24.77} &
  22.35 \\ \hline
 &
  Toxicity &
  \multicolumn{1}{c|}{37.36} &
  \multicolumn{1}{c|}{47.63} &
  10.67 &
  \multicolumn{1}{c|}{4.32} &
  \multicolumn{1}{c|}{47.25} &
  4.68 \\ \cline{2-8} 
\multirow{-2}{*}{GPT2 1.61B} &
  PPL (WikiText2) &
  \multicolumn{1}{c|}{18.50} &
  \multicolumn{1}{c|}{20.95} &
  18.16 &
  \multicolumn{1}{c|}{18.80} &
  \multicolumn{1}{c|}{18.01} &
  18.15 \\ \hline
 &
  Toxicity &
  \multicolumn{1}{c|}{39.71} &
  \multicolumn{1}{c|}{38.79} &
  5.69 &
  \multicolumn{1}{c|}{40.57} &
  \multicolumn{1}{c|}{40.86} &
  4.94 \\ \cline{2-8} 
\multirow{-2}{*}{GPT Neo 1.3B} &
  PPL (WikiText2) &
  \multicolumn{1}{c|}{14.85} &
  \multicolumn{1}{c|}{14.98} &
  16.56 &
  \multicolumn{1}{c|}{14.86} &
  \multicolumn{1}{c|}{15.01} &
  16.52 \\ \hline
\rowcolor[HTML]{EFEFEF} 
\cellcolor[HTML]{EFEFEF} &
  Toxicity &
  \multicolumn{1}{c|}{\cellcolor[HTML]{EFEFEF}40.87} &
  \multicolumn{1}{c|}{\cellcolor[HTML]{EFEFEF}41.08} &
  8.03 &
  \multicolumn{1}{c|}{\cellcolor[HTML]{EFEFEF}27.98} &
  \multicolumn{1}{c|}{\cellcolor[HTML]{EFEFEF}43.40} &
  4.91 \\ \cline{2-8} 
\rowcolor[HTML]{EFEFEF} 
\multirow{-2}{*}{\cellcolor[HTML]{EFEFEF}\textbf{Overall}} &
  PPL (WikiText2) &
  \multicolumn{1}{c|}{\cellcolor[HTML]{EFEFEF}18.04} &
  \multicolumn{1}{c|}{\cellcolor[HTML]{EFEFEF}19.18} &
  18.55 &
  \multicolumn{1}{c|}{\cellcolor[HTML]{EFEFEF}18.85} &
  \multicolumn{1}{c|}{\cellcolor[HTML]{EFEFEF}19.26} &
  19.01 \\ \hline
\end{tabular}
\label{tab:rq3}
\end{table}
\subsection{RQ3: Does the Dynamic Selection Employed by \nick Offer Any Advantage?}
\label{sec:rq3}

In the final research question, we investigate the effectiveness of dynamic slicing in delivering focused model repair. As described in \S~\ref{sec:baseline}, we evaluated two additional variants of both the standard \nick and \nickkl: \nick\textit{+ Min} and \nick\textit{+ Fixed}. The \nick\textit{+ Min} variant selects the transformer block with the lowest error concentration, as opposed to the highest in the regular \nick. This baseline allows us to assess the impact of selection on repair efficacy. Similarly, \nick\textit{+ Fixed} disables dynamic slicing and instead pre-selects the block with the highest error concentration for repair. This variant enables us to assess the impact of dynamic selection on repair effectiveness. The impact on model performance in this RQ is evaluated solely on the \textit{WikiText2} benchmark.

\begin{figure}
\centering
\begin{subfigure}{.4\textwidth}
  \centering
  \includegraphics[width=1\linewidth]{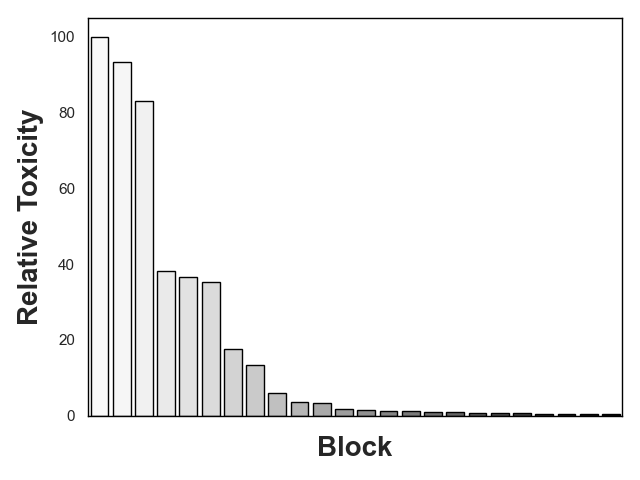}
  \caption{GPT Neo 1.3B}
  \label{fig:gpt2xl}
\end{subfigure}%
\begin{subfigure}{.4\textwidth}
  \centering
  \includegraphics[width=1\linewidth]{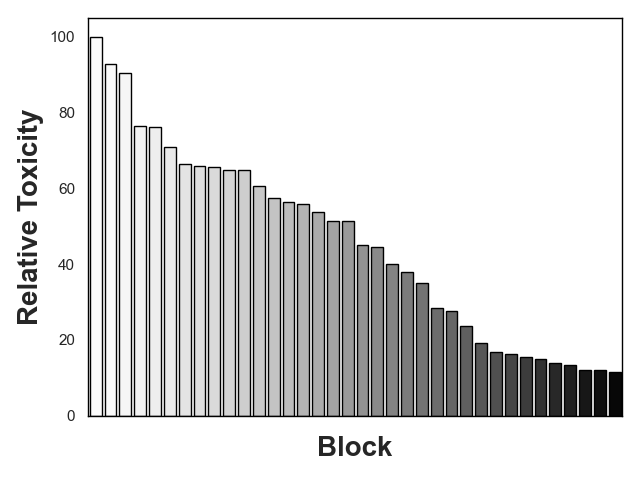}
  \caption{GPT2 812M}
  \label{fig:gpt2large}
\end{subfigure}
\begin{subfigure}{.4\textwidth}
  \centering
  \includegraphics[width=1\linewidth]{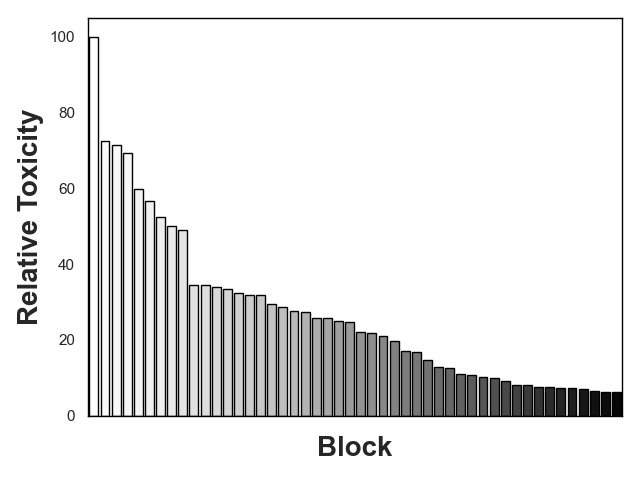}
  \caption{GPT2 1.61B}
  \label{fig:gptneo}
\end{subfigure}
\caption{Relative Toxicity Levels of Transformer Blocks Across Different Models}
\label{fig:tox}
\end{figure}

Table~\ref{tab:rq3} presents the comparative results of these variants against the regular \nick. It demonstrates that both regular \nick variants significantly outperform their \textit{Min} and \textit{Fixed} counterparts. For instance, standard \nick and \nickkl reduce toxicity by 80.5\% and 88.6\% more than their \textit{Min} variants while maintaining a similar level of PPL (with regular \textit{IRepairs} showing 3.3\% and 1.3\% more reduction, respectively). It clearly shows the impact of selection made by regular \nick in repairing the model. 

Similarly, both regular \textit{IRepairs} outperform their \textit{Fixed} counterparts by a clear margin. Standard \nick and \nickkl reduce toxicity by 80.4\% and 82.5\% more than their \textit{Fixed} counterparts while achieving slightly lower PPL (with \textit{Fixed} variants scoring 2.7\% and 0.8\% less in PPL). Additionally, pre-selecting the most error-prone blocks and focusing repair efforts on them slightly outperforms the dynamic \textit{Min} variants, with the difference being more noticeable in the \textit{Fixed + KL} variant (scoring 35\% less in toxicity than \textit{Min}).

We particularly observed that the \textit{Fixed + KL} approach performs comparably to the \nickkl method on the GPT2 1.61B model. To understand why fixed selection was effective for this model, we analyzed the toxicity levels of all transformer blocks across different models. We calculated toxicity by randomly sampling 2000 examples and computing the average sensitivity for each block. Figure~\ref{fig:tox} displays the distribution of relative toxicity across all blocks for the models studied.

For the GPT2 1.61B model, the most toxic block is about 28\% more toxic than the second most toxic block. In contrast, for the GPT2 812M model, the difference between the most and second most toxic blocks is only 8\% and 9.5\% with the third most toxic blocks. Similarly, for the GPT Neo 1.3B model, the difference between the most and second most toxic blocks is just 6.5\%. This indicates that the GPT2 1.61B model has a higher concentration of toxicity in the top block, making fixed targeted repair of this block more effective. In other models, errors are more evenly distributed among the top few blocks, reducing the effectiveness of a top-block-only repair.

This suggests that errors can be dispersed throughout the model, and a fixed selection technique may require costly tuning to determine the optimal selection threshold. In contrast, our dynamic slicing approach avoids this need for tuning and allows for model-wide repair by dynamically focusing on the most error-prone areas during training.

Figure~\ref{fig:tox} also shows that GPT-2 1.61B, GPT-2 812M, and GPT-Neo 1.3B have 245.3\%, 120.8\%, and 1137.7\% higher average error density in the top 20\% of blocks compared to the remaining 80\% of blocks. Error density was measured by dividing the total toxicity within \( N \) blocks by \( N \). This clearly indicates that the repair process should give more priority to these highly error-inducing regions than to others, which may lead to superior outcomes, as observed in our results.

\section{Related Work}
Software engineering research has proposed various techniques to repair bugs in deep neural networks (DNNs) that arise during training or within the network structure itself~\cite{zhang2019apricot,zhang2021autotrainer,wardat2022deepdiagnosis,ma2018mode}. Examples include Zhang \textit{et al.} 's method for monitoring DNN training and suggesting corrective actions for anomalies~\cite{zhang2021autotrainer}, and Wardat \textit{et al.}'s work on identifying and fixing structural bugs in DNNs~\cite{wardat2022deepdiagnosis}. However, these techniques primarily address issues stemming from the DNN itself. In contrast, data-driven errors in LLMs, such as toxicity or hallucinations, can stem from biases and inconsistencies within the training data itself~\cite{ji2023survey}, requiring solutions beyond structural or training bug fixes.

On the other hand, machine learning research offers several strategies to mitigate such errors, primarily operating in three stages~\cite{wang2022exploring,korbak2023pretraining,pan2023automatically}: runtime methods~\cite{liu2021dexperts,dathathri2019plug,wang2022exploring,schick2021self,leong2023self,yang2022unified,xu2022leashing,niu2024parameter,weng2023large},  the pre-training stage~\cite{korbak2023pretraining,liu2024exposing,huang2023survey}, and alignment stage~\cite{wang2022exploring,gehman2020realtoxicityprompts,lee2024mechanistic,liu2023chain,rafailov2024direct}. These approaches are often complementary, each with its own scope and applicability~\cite{huang2023survey}. 
Runtime methods aim to control model outputs through non-invasive techniques such as context augmentation, vocabulary modification, word filtering, and output post-processing~\cite{dathathri2019plug,wang2022exploring,huang2023survey}. These methods provide flexible runtime control, though they face certain constraints - they do not modify the underlying model architecture or weights when such modifications would be beneficial~\cite{wang2022exploring}. The additional computational overhead also presents challenges in latency-sensitive applications~\cite{wang2022exploring}. Prompt-based approaches for steering instruction-following models fall into this category as well, relying on prompt engineering techniques to mitigate undesirable responses ~\cite{ganguli2023capacity, xie2023defending}.

In contrast to runtime methods, pretraining-based approaches suggest removing problematic data from the training corpus or modifying model architecture~\cite{korbak2023pretraining,
liu2024exposing}. While effective, this can be prohibitively expensive, making them suitable for new model developments~\cite{wang2022exploring}. DAT methods that operate during the alignment stage, like ours, offer a preemptive approach to error mitigation. This is particularly suitable in scenarios where these limitations are especially impactful and require a more invasive repair~\cite{wang2022exploring, lee2024mechanistic}.

DAT methods aim to continue pretraining or fine-tuning models on domain-specific texts~\cite{gehman2020realtoxicityprompts,lee2024mechanistic}. For instance, Gehman \textit{et al.} \cite{gehman2020realtoxicityprompts} applied a framework called Domain-Adaptive Pretraining (DAPT)~\cite{gururangan2020don} to further pretrain GPT-2 on curated non-toxic data, reducing its toxicity. Similarly, Wang \textit{et al.} \cite{wang2022exploring} used the DAPT framework with self-generated data to detoxify models. However, their technique is model-dependent and relies on self-generated data, which may not be applicable in all error scenarios, such as factual inaccuracies or hallucinations. Solaiman and Dennison~\cite{solaiman2021process} proposed a general framework for aligning language models to specific target values, but it involves expensive iterative training~\cite{si2022prompting}. These methods aim to optimize pre-trained model parameters using domain-specific texts, mitigating errors while minimizing the impact on overall performance.

Reinforcement learning (RL) from human feedback (RLHF) is another paradigm within DAT methods that relies on human demonstration datasets to align language models, and it has been shown to mitigate errors in LMs~\cite{ouyang2022training}. The recently proposed Direct Preference Optimization (DPO) is a cutting-edge RL-inspired algorithm designed to overcome the instability and complexity of RLHF~\cite{rafailov2024direct}. In recent work, Lee \textit{et al.} \cite{lee2024mechanistic} demonstrated the effectiveness of the DPO framework in detoxifying the GPT-2 model using paired datasets of toxic and non-toxic examples.

However, these existing domain-adaptive techniques treat all model parameters uniformly during repair. This indiscriminate approach increases the risk of altering parameters unrelated to the specific errors being addressed, which can disrupt the general knowledge stored in those parameters. Such an ``intent-unaware'' approach not only risks harming overall performance but is also limited in effectively targeting the error-prone parts of the model. A more focused strategy could address these issues more efficiently by concentrating the repair effort where it is most needed. Our proposed technique, \nick, addresses these concerns by enabling a selective repair strategy.

Parameter-efficient fine-tuning (PEFT)~\cite{ding2023parameter} refers to a family of model alignment methods, such as LoRA~\cite{hu2021lora}, designed to optimize resource and computational efficiency during fine-tuning. In contrast, \nick emphasizes targeted adjustments to address specific errors in a model’s behavior. Like existing repair methods, intent awareness is not an explicit goal of PEFT methods either.

Knowledge editing (KE) is a related area within machine learning that focuses on updating a model's factual knowledge, allowing developers and end-users to modify the model beyond the training setup~\cite{mitchell2021fast,meng2022locating,wang2024detoxifying}. These techniques complement training-time methods by enabling model fixes during test time~\cite{mitchell2021fast}. While training-time methods aim for global correction, KE techniques provide localized fixes by updating the model's knowledge with a single instance~\cite{mitchell2021fast,meng2022locating,wang2024detoxifying}.

\section{Threats To Validity and Limitations}
\label{sec:threats}
An internal threat to the study is the quality of the detoxification and evaluation datasets. To address this, we utilize both the training dataset and evaluation setup from a recent reputable work on LLM detoxification~\cite{lee2024mechanistic}. Additionally, we employ the implementations provided in the same study to address concerns about the construct validity of baseline techniques. Another internal threat arises from the reliability of the evaluation metrics. To mitigate this concern, we measure repair or toxicity scores using the widely used Perspective API~\cite{papi} and evaluate model quality post-repair using perplexity, as employed in many prior works~\cite{lee2024mechanistic, wang2022exploring, gehman2020realtoxicityprompts}. Similarly, our evaluation metrics for measuring computational overheads are based on well-established metrics in the literature~\cite{kaplan2020scaling, lee2019energy}. An external threat is the relevance of the models used. To address this, we have selected three models from the GPT and GPT-Neo families with billions of parameters, all of which have previously been employed for evaluating detoxification techniques~\cite{leong2023self, gehman2020realtoxicityprompts, xu2022leashing,korbak2023pretraining, lee2024mechanistic, yang2022unified}.

While the case study performed in this paper shows that \nick can effectively address the data-driven errors in large language models (LLMs), it is demonstrated within the context of detoxification. Further research is encouraged to explore its effectiveness and generalizability to other data-driven error scenarios, which will enhance the understanding and potential applications of this approach. Furthermore, although we evaluated \nick on models with billions of parameters—similar to those frequently used in evaluating prior repair techniques—its performance in ultra-large-scale LLMs remains an area for further investigation.

\section{Conclusion}
\label{sec:conclusion}

In this paper, we introduce \nick, an intent-aware technique for selectively repairing data-driven errors in LLMs through dynamic model slicing. While domain-adaptive training with curated data has shown promise, it tends to optimize model parameters indiscriminately, which can limit repair efficacy and increase the risk of negatively affecting general model performance by altering unrelated parameters. To address these limitations, \nick identifies the relevant portions of the model responsible for the errors, allowing for more targeted repair and making it an intent-aware approach. Our method employs a gradient-based technique to select the most relevant parts of the model by analyzing sensitivity to slicing criteria. Unlike existing slicing routines, our technique is specifically designed to address transformer-related challenges and to avoid the need for expensive tuning of selection thresholds. In a case study focused on model detoxification, \nick demonstrated its effectiveness in addressing the root causes of toxicity while minimizing the impact on general performance, outperforming state-of-the-art baselines. Our empirical results also suggest that errors can be highly concentrated in very limited regions of the model, highlighting the need for selective repair. We further demonstrate that a dynamic selection-based repair strategy is essential for effectively addressing errors dispersed throughout the model.

\section{Data Availability Statement}
The replication package is available here~\cite{irepair2024} and includes all the results, code, and data, along with a `readme' file that provides detailed instructions on how to reproduce the results.

\section*{Acknowledgment}
This work was supported in part by US NSF grants 2512857 and 2512858. We want to thank the reviewers for their valuable and insightful comments. The views expressed in this work are solely those of the authors and do not reflect the opinions of the sponsors.


\bibliographystyle{ACM-Reference-Format}
\bibliography{refs.bib}


\begin{thebibliography}{60}


\ifx \showCODEN    \undefined \def \showCODEN     #1{\unskip}     \fi
\ifx \showISBNx    \undefined \def \showISBNx     #1{\unskip}     \fi
\ifx \showISBNxiii \undefined \def \showISBNxiii  #1{\unskip}     \fi
\ifx \showISSN     \undefined \def \showISSN      #1{\unskip}     \fi
\ifx \showLCCN     \undefined \def \showLCCN      #1{\unskip}     \fi
\ifx \shownote     \undefined \def \shownote      #1{#1}          \fi
\ifx \showarticletitle \undefined \def \showarticletitle #1{#1}   \fi
\ifx \showURL      \undefined \def \showURL       {\relax}        \fi
\providecommand\bibfield[2]{#2}
\providecommand\bibinfo[2]{#2}
\providecommand\natexlab[1]{#1}
\providecommand\showeprint[2][]{arXiv:#2}

\bibitem[gpt(2024a)]%
        {gptneo}
 \bibinfo{year}{2024}\natexlab{a}.
\newblock \bibinfo{booktitle}{\emph{GPT Neo Models by Eleuther AI}}.
\newblock
\urldef\tempurl%
\url{https://www.eleuther.ai/artifacts/gpt-neo}
\showURL{%
Retrieved August 31, 2024 from \tempurl}


\bibitem[gpt(2024b)]%
        {gptneohf}
 \bibinfo{year}{2024}\natexlab{b}.
\newblock \bibinfo{booktitle}{\emph{GPT Neo Models by Eleuther AI: HuggingFace Repository}}.
\newblock
\urldef\tempurl%
\url{https://huggingface.co/EleutherAI/}
\showURL{%
Retrieved August 31, 2024 from \tempurl}


\bibitem[gpt(2024c)]%
        {gpthf}
 \bibinfo{year}{2024}\natexlab{c}.
\newblock \bibinfo{booktitle}{\emph{GPT2 Models by OpenAI: HuggingFace Repository}}.
\newblock
\urldef\tempurl%
\url{https://huggingface.co/openai}
\showURL{%
Retrieved August 31, 2024 from \tempurl}


\bibitem[pap(2024)]%
        {papi}
 \bibinfo{year}{2024}\natexlab{}.
\newblock \bibinfo{booktitle}{\emph{Perspective API}}.
\newblock
\urldef\tempurl%
\url{https://perspectiveapi.com/}
\showURL{%
Retrieved August 31, 2024 from \tempurl}


\bibitem[tor(2024)]%
        {torchcuda}
 \bibinfo{year}{2024}\natexlab{}.
\newblock \bibinfo{booktitle}{\emph{PyTorch CUDA API}}.
\newblock
\urldef\tempurl%
\url{https://pytorch.org/docs/stable/cuda.html}
\showURL{%
Retrieved August 31, 2024 from \tempurl}


\bibitem[Casson(2023)]%
        {casson2023transformerflops}
\bibfield{author}{\bibinfo{person}{Adam Casson}.} \bibinfo{year}{2023}\natexlab{}.
\newblock \showarticletitle{Transformer FLOPs}.
\newblock  (\bibinfo{year}{2023}).
\newblock
\urldef\tempurl%
\url{https://adamcasson.com/posts/transformer-flops}
\showURL{%
\tempurl}


\bibitem[Csisz{\'a}r(1975)]%
        {csiszar1975divergence}
\bibfield{author}{\bibinfo{person}{Imre Csisz{\'a}r}.} \bibinfo{year}{1975}\natexlab{}.
\newblock \showarticletitle{I-divergence geometry of probability distributions and minimization problems}.
\newblock \bibinfo{journal}{\emph{The annals of probability}} (\bibinfo{year}{1975}), \bibinfo{pages}{146--158}.
\newblock


\bibitem[Dathathri et~al\mbox{.}(2020)]%
        {dathathri2019plug}
\bibfield{author}{\bibinfo{person}{Sumanth Dathathri}, \bibinfo{person}{Andrea Madotto}, \bibinfo{person}{Janice Lan}, \bibinfo{person}{Jane Hung}, \bibinfo{person}{Eric Frank}, \bibinfo{person}{Piero Molino}, \bibinfo{person}{Jason Yosinski}, {and} \bibinfo{person}{Rosanne Liu}.} \bibinfo{year}{2020}\natexlab{}.
\newblock \showarticletitle{Plug and Play Language Models: A Simple Approach to Controlled Text Generation}. In \bibinfo{booktitle}{\emph{International Conference on Learning Representations}}.
\newblock
\urldef\tempurl%
\url{https://openreview.net/forum?id=H1edEyBKDS}
\showURL{%
\tempurl}


\bibitem[Ding et~al\mbox{.}(2022)]%
        {ding2022delta}
\bibfield{author}{\bibinfo{person}{Ning Ding}, \bibinfo{person}{Yujia Qin}, \bibinfo{person}{Guang Yang}, \bibinfo{person}{Fuchao Wei}, \bibinfo{person}{Zonghan Yang}, \bibinfo{person}{Yusheng Su}, \bibinfo{person}{Shengding Hu}, \bibinfo{person}{Yulin Chen}, \bibinfo{person}{Chi-Min Chan}, \bibinfo{person}{Weize Chen}, {et~al\mbox{.}}} \bibinfo{year}{2022}\natexlab{}.
\newblock \showarticletitle{Delta tuning: A comprehensive study of parameter efficient methods for pre-trained language models}.
\newblock \bibinfo{journal}{\emph{arXiv preprint arXiv:2203.06904}} (\bibinfo{year}{2022}).
\newblock
\href{https://doi.org/10.48550/arXiv.2203.06904}{doi:\nolinkurl{10.48550/arXiv.2203.06904}}


\bibitem[Ding et~al\mbox{.}(2023)]%
        {ding2023parameter}
\bibfield{author}{\bibinfo{person}{Ning Ding}, \bibinfo{person}{Yujia Qin}, \bibinfo{person}{Guang Yang}, \bibinfo{person}{Fuchao Wei}, \bibinfo{person}{Zonghan Yang}, \bibinfo{person}{Yusheng Su}, \bibinfo{person}{Shengding Hu}, \bibinfo{person}{Yulin Chen}, \bibinfo{person}{Chi-Min Chan}, \bibinfo{person}{Weize Chen}, {et~al\mbox{.}}} \bibinfo{year}{2023}\natexlab{}.
\newblock \showarticletitle{Parameter-efficient fine-tuning of large-scale pre-trained language models}.
\newblock \bibinfo{journal}{\emph{Nature Machine Intelligence}} \bibinfo{volume}{5}, \bibinfo{number}{3} (\bibinfo{year}{2023}), \bibinfo{pages}{220--235}.
\newblock
\href{https://doi.org/10.1038/s42256-023-00626-4}{doi:\nolinkurl{10.1038/s42256-023-00626-4}}


\bibitem[Ganguli et~al\mbox{.}(2023)]%
        {ganguli2023capacity}
\bibfield{author}{\bibinfo{person}{Deep Ganguli}, \bibinfo{person}{Amanda Askell}, \bibinfo{person}{Nicholas Schiefer}, \bibinfo{person}{Thomas~I Liao}, \bibinfo{person}{Kamil{\.e} Luko{\v{s}}i{\=u}t{\.e}}, \bibinfo{person}{Anna Chen}, \bibinfo{person}{Anna Goldie}, \bibinfo{person}{Azalia Mirhoseini}, \bibinfo{person}{Catherine Olsson}, \bibinfo{person}{Danny Hernandez}, {et~al\mbox{.}}} \bibinfo{year}{2023}\natexlab{}.
\newblock \showarticletitle{The capacity for moral self-correction in large language models}.
\newblock \bibinfo{journal}{\emph{arXiv preprint arXiv:2302.07459}} (\bibinfo{year}{2023}).
\newblock
\href{https://doi.org/10.48550/arXiv.2302.07459}{doi:\nolinkurl{10.48550/arXiv.2302.07459}}


\bibitem[Gehman et~al\mbox{.}(2020)]%
        {gehman2020realtoxicityprompts}
\bibfield{author}{\bibinfo{person}{Samuel Gehman}, \bibinfo{person}{Suchin Gururangan}, \bibinfo{person}{Maarten Sap}, \bibinfo{person}{Yejin Choi}, {and} \bibinfo{person}{Noah~A. Smith}.} \bibinfo{year}{2020}\natexlab{}.
\newblock \showarticletitle{{R}eal{T}oxicity{P}rompts: Evaluating Neural Toxic Degeneration in Language Models}. In \bibinfo{booktitle}{\emph{Findings of the Association for Computational Linguistics: EMNLP 2020}}, \bibfield{editor}{\bibinfo{person}{Trevor Cohn}, \bibinfo{person}{Yulan He}, {and} \bibinfo{person}{Yang Liu}} (Eds.). \bibinfo{publisher}{Association for Computational Linguistics}, \bibinfo{address}{Online}, \bibinfo{pages}{3356--3369}.
\newblock
\href{https://doi.org/10.18653/v1/2020.findings-emnlp.301}{doi:\nolinkurl{10.18653/v1/2020.findings-emnlp.301}}


\bibitem[Geva et~al\mbox{.}(2022)]%
        {geva2022transformer}
\bibfield{author}{\bibinfo{person}{Mor Geva}, \bibinfo{person}{Avi Caciularu}, \bibinfo{person}{Kevin Wang}, {and} \bibinfo{person}{Yoav Goldberg}.} \bibinfo{year}{2022}\natexlab{}.
\newblock \showarticletitle{Transformer Feed-Forward Layers Build Predictions by Promoting Concepts in the Vocabulary Space}. In \bibinfo{booktitle}{\emph{Proceedings of the 2022 Conference on Empirical Methods in Natural Language Processing}}, \bibfield{editor}{\bibinfo{person}{Yoav Goldberg}, \bibinfo{person}{Zornitsa Kozareva}, {and} \bibinfo{person}{Yue Zhang}} (Eds.). \bibinfo{publisher}{Association for Computational Linguistics}, \bibinfo{address}{Abu Dhabi, United Arab Emirates}, \bibinfo{pages}{30--45}.
\newblock
\href{https://doi.org/10.18653/v1/2022.emnlp-main.3}{doi:\nolinkurl{10.18653/v1/2022.emnlp-main.3}}


\bibitem[Gururangan et~al\mbox{.}(2020)]%
        {gururangan2020don}
\bibfield{author}{\bibinfo{person}{Suchin Gururangan}, \bibinfo{person}{Ana Marasovi{\'c}}, \bibinfo{person}{Swabha Swayamdipta}, \bibinfo{person}{Kyle Lo}, \bibinfo{person}{Iz Beltagy}, \bibinfo{person}{Doug Downey}, {and} \bibinfo{person}{Noah~A. Smith}.} \bibinfo{year}{2020}\natexlab{}.
\newblock \showarticletitle{Don`t Stop Pretraining: Adapt Language Models to Domains and Tasks}. In \bibinfo{booktitle}{\emph{Proceedings of the 58th Annual Meeting of the Association for Computational Linguistics}}, \bibfield{editor}{\bibinfo{person}{Dan Jurafsky}, \bibinfo{person}{Joyce Chai}, \bibinfo{person}{Natalie Schluter}, {and} \bibinfo{person}{Joel Tetreault}} (Eds.). \bibinfo{publisher}{Association for Computational Linguistics}, \bibinfo{address}{Online}, \bibinfo{pages}{8342--8360}.
\newblock
\href{https://doi.org/10.18653/v1/2020.acl-main.740}{doi:\nolinkurl{10.18653/v1/2020.acl-main.740}}


\bibitem[Gyim{\'o}thy et~al\mbox{.}(1999)]%
        {gyimothy1999efficient}
\bibfield{author}{\bibinfo{person}{Tibor Gyim{\'o}thy}, \bibinfo{person}{Arp{\'a}d Besz{\'e}des}, {and} \bibinfo{person}{Ist{\'a}n Forg{\'a}cs}.} \bibinfo{year}{1999}\natexlab{}.
\newblock \showarticletitle{An efficient relevant slicing method for debugging}.
\newblock \bibinfo{journal}{\emph{ACM SIGSOFT Software Engineering Notes}} \bibinfo{volume}{24}, \bibinfo{number}{6} (\bibinfo{year}{1999}), \bibinfo{pages}{303--321}.
\newblock
\href{https://doi.org/10.1145/318774.319248}{doi:\nolinkurl{10.1145/318774.319248}}


\bibitem[Hu et~al\mbox{.}(2021)]%
        {hu2021lora}
\bibfield{author}{\bibinfo{person}{Edward~J Hu}, \bibinfo{person}{Yelong Shen}, \bibinfo{person}{Phillip Wallis}, \bibinfo{person}{Zeyuan Allen-Zhu}, \bibinfo{person}{Yuanzhi Li}, \bibinfo{person}{Shean Wang}, \bibinfo{person}{Lu Wang}, {and} \bibinfo{person}{Weizhu Chen}.} \bibinfo{year}{2021}\natexlab{}.
\newblock \showarticletitle{Lora: Low-rank adaptation of large language models}.
\newblock \bibinfo{journal}{\emph{arXiv preprint arXiv:2106.09685}} (\bibinfo{year}{2021}).
\newblock
\href{https://doi.org/10.48550/arXiv.2106.09685}{doi:\nolinkurl{10.48550/arXiv.2106.09685}}


\bibitem[Huang et~al\mbox{.}(2023)]%
        {huang2023survey}
\bibfield{author}{\bibinfo{person}{Lei Huang}, \bibinfo{person}{Weijiang Yu}, \bibinfo{person}{Weitao Ma}, \bibinfo{person}{Weihong Zhong}, \bibinfo{person}{Zhangyin Feng}, \bibinfo{person}{Haotian Wang}, \bibinfo{person}{Qianglong Chen}, \bibinfo{person}{Weihua Peng}, \bibinfo{person}{Xiaocheng Feng}, \bibinfo{person}{Bing Qin}, {et~al\mbox{.}}} \bibinfo{year}{2023}\natexlab{}.
\newblock \showarticletitle{A survey on hallucination in large language models: Principles, taxonomy, challenges, and open questions}.
\newblock \bibinfo{journal}{\emph{ACM Transactions on Information Systems}} (\bibinfo{year}{2023}).
\newblock
\href{https://doi.org/10.1145/3703155}{doi:\nolinkurl{10.1145/3703155}}


\bibitem[Imtiaz et~al\mbox{.}(2024)]%
        {irepair2024}
\bibfield{author}{\bibinfo{person}{Sayem~Mohammad Imtiaz}, \bibinfo{person}{Astha Singh}, \bibinfo{person}{Fraol Batole}, {and} \bibinfo{person}{Hridesh Rajan}.} \bibinfo{year}{2024}\natexlab{}.
\newblock \bibinfo{booktitle}{\emph{IRepair - results and replication package}}.
\newblock
\urldef\tempurl%
\url{https://huggingface.co/datasets/Anonymous007/IRepair/tree/main}
\showURL{%
Retrieved August 31, 2024 from \tempurl}


\bibitem[Ji et~al\mbox{.}(2023)]%
        {ji2023survey}
\bibfield{author}{\bibinfo{person}{Ziwei Ji}, \bibinfo{person}{Nayeon Lee}, \bibinfo{person}{Rita Frieske}, \bibinfo{person}{Tiezheng Yu}, \bibinfo{person}{Dan Su}, \bibinfo{person}{Yan Xu}, \bibinfo{person}{Etsuko Ishii}, \bibinfo{person}{Ye~Jin Bang}, \bibinfo{person}{Andrea Madotto}, {and} \bibinfo{person}{Pascale Fung}.} \bibinfo{year}{2023}\natexlab{}.
\newblock \showarticletitle{Survey of hallucination in natural language generation}.
\newblock \bibinfo{journal}{\emph{Comput. Surveys}} \bibinfo{volume}{55}, \bibinfo{number}{12} (\bibinfo{year}{2023}), \bibinfo{pages}{1--38}.
\newblock
\href{https://doi.org/10.1145/3571730}{doi:\nolinkurl{10.1145/3571730}}


\bibitem[Kaplan et~al\mbox{.}(2020)]%
        {kaplan2020scaling}
\bibfield{author}{\bibinfo{person}{Jared Kaplan}, \bibinfo{person}{Sam McCandlish}, \bibinfo{person}{Tom Henighan}, \bibinfo{person}{Tom~B Brown}, \bibinfo{person}{Benjamin Chess}, \bibinfo{person}{Rewon Child}, \bibinfo{person}{Scott Gray}, \bibinfo{person}{Alec Radford}, \bibinfo{person}{Jeffrey Wu}, {and} \bibinfo{person}{Dario Amodei}.} \bibinfo{year}{2020}\natexlab{}.
\newblock \showarticletitle{Scaling laws for neural language models}.
\newblock \bibinfo{journal}{\emph{arXiv preprint arXiv:2001.08361}} (\bibinfo{year}{2020}).
\newblock
\href{https://doi.org/10.48550/arXiv.2001.08361}{doi:\nolinkurl{10.48550/arXiv.2001.08361}}


\bibitem[Kirkpatrick et~al\mbox{.}(2017)]%
        {kirkpatrick2017overcoming}
\bibfield{author}{\bibinfo{person}{James Kirkpatrick}, \bibinfo{person}{Razvan Pascanu}, \bibinfo{person}{Neil Rabinowitz}, \bibinfo{person}{Joel Veness}, \bibinfo{person}{Guillaume Desjardins}, \bibinfo{person}{Andrei~A Rusu}, \bibinfo{person}{Kieran Milan}, \bibinfo{person}{John Quan}, \bibinfo{person}{Tiago Ramalho}, \bibinfo{person}{Agnieszka Grabska-Barwinska}, {et~al\mbox{.}}} \bibinfo{year}{2017}\natexlab{}.
\newblock \showarticletitle{Overcoming catastrophic forgetting in neural networks}.
\newblock \bibinfo{journal}{\emph{Proceedings of the national academy of sciences}} \bibinfo{volume}{114}, \bibinfo{number}{13} (\bibinfo{year}{2017}), \bibinfo{pages}{3521--3526}.
\newblock
\href{https://doi.org/10.1073/pnas.1611835114}{doi:\nolinkurl{10.1073/pnas.1611835114}}


\bibitem[Korbak et~al\mbox{.}(2023)]%
        {korbak2023pretraining}
\bibfield{author}{\bibinfo{person}{Tomasz Korbak}, \bibinfo{person}{Kejian Shi}, \bibinfo{person}{Angelica Chen}, \bibinfo{person}{Rasika~Vinayak Bhalerao}, \bibinfo{person}{Christopher Buckley}, \bibinfo{person}{Jason Phang}, \bibinfo{person}{Samuel~R Bowman}, {and} \bibinfo{person}{Ethan Perez}.} \bibinfo{year}{2023}\natexlab{}.
\newblock \showarticletitle{Pretraining language models with human preferences}. In \bibinfo{booktitle}{\emph{International Conference on Machine Learning}}. PMLR, \bibinfo{pages}{17506--17533}.
\newblock


\bibitem[Lee et~al\mbox{.}(2024)]%
        {lee2024mechanistic}
\bibfield{author}{\bibinfo{person}{Andrew Lee}, \bibinfo{person}{Xiaoyan Bai}, \bibinfo{person}{Itamar Pres}, \bibinfo{person}{Martin Wattenberg}, \bibinfo{person}{Jonathan~K. Kummerfeld}, {and} \bibinfo{person}{Rada Mihalcea}.} \bibinfo{year}{2024}\natexlab{}.
\newblock \showarticletitle{A mechanistic understanding of alignment algorithms: a case study on DPO and toxicity}. In \bibinfo{booktitle}{\emph{Proceedings of the 41st International Conference on Machine Learning}} (Vienna, Austria) \emph{(\bibinfo{series}{ICML'24})}. \bibinfo{publisher}{JMLR.org}, Article \bibinfo{articleno}{1052}, \bibinfo{numpages}{18}~pages.
\newblock


\bibitem[Lee et~al\mbox{.}(2019)]%
        {lee2019energy}
\bibfield{author}{\bibinfo{person}{Youngwan Lee}, \bibinfo{person}{Joong-won Hwang}, \bibinfo{person}{Sangrok Lee}, \bibinfo{person}{Yuseok Bae}, {and} \bibinfo{person}{Jongyoul Park}.} \bibinfo{year}{2019}\natexlab{}.
\newblock \showarticletitle{An energy and GPU-computation efficient backbone network for real-time object detection}. In \bibinfo{booktitle}{\emph{Proceedings of the IEEE/CVF conference on computer vision and pattern recognition workshops}}. \bibinfo{pages}{0--0}.
\newblock


\bibitem[Leong et~al\mbox{.}(2023)]%
        {leong2023self}
\bibfield{author}{\bibinfo{person}{Chak~Tou Leong}, \bibinfo{person}{Yi Cheng}, \bibinfo{person}{Jiashuo Wang}, \bibinfo{person}{Jian Wang}, {and} \bibinfo{person}{Wenjie Li}.} \bibinfo{year}{2023}\natexlab{}.
\newblock \showarticletitle{Self-Detoxifying Language Models via Toxification Reversal}. In \bibinfo{booktitle}{\emph{Proceedings of the 2023 Conference on Empirical Methods in Natural Language Processing}}, \bibfield{editor}{\bibinfo{person}{Houda Bouamor}, \bibinfo{person}{Juan Pino}, {and} \bibinfo{person}{Kalika Bali}} (Eds.). \bibinfo{publisher}{Association for Computational Linguistics}, \bibinfo{address}{Singapore}, \bibinfo{pages}{4433--4449}.
\newblock
\href{https://doi.org/10.18653/v1/2023.emnlp-main.269}{doi:\nolinkurl{10.18653/v1/2023.emnlp-main.269}}


\bibitem[Liu et~al\mbox{.}(2021)]%
        {liu2021dexperts}
\bibfield{author}{\bibinfo{person}{Alisa Liu}, \bibinfo{person}{Maarten Sap}, \bibinfo{person}{Ximing Lu}, \bibinfo{person}{Swabha Swayamdipta}, \bibinfo{person}{Chandra Bhagavatula}, \bibinfo{person}{Noah~A Smith}, {and} \bibinfo{person}{Yejin Choi}.} \bibinfo{year}{2021}\natexlab{}.
\newblock \showarticletitle{DExperts: Decoding-time controlled text generation with experts and anti-experts}.
\newblock \bibinfo{journal}{\emph{arXiv preprint arXiv:2105.03023}} (\bibinfo{year}{2021}).
\newblock
\href{https://doi.org/10.48550/arXiv.2105.03023}{doi:\nolinkurl{10.48550/arXiv.2105.03023}}


\bibitem[Liu et~al\mbox{.}(2024)]%
        {liu2024exposing}
\bibfield{author}{\bibinfo{person}{Bingbin Liu}, \bibinfo{person}{Jordan Ash}, \bibinfo{person}{Surbhi Goel}, \bibinfo{person}{Akshay Krishnamurthy}, {and} \bibinfo{person}{Cyril Zhang}.} \bibinfo{year}{2024}\natexlab{}.
\newblock \showarticletitle{Exposing attention glitches with flip-flop language modeling}.
\newblock \bibinfo{journal}{\emph{Advances in Neural Information Processing Systems}}  \bibinfo{volume}{36} (\bibinfo{year}{2024}).
\newblock


\bibitem[Liu et~al\mbox{.}(2023)]%
        {liu2023chain}
\bibfield{author}{\bibinfo{person}{Hao Liu}, \bibinfo{person}{Carmelo Sferrazza}, {and} \bibinfo{person}{Pieter Abbeel}.} \bibinfo{year}{2023}\natexlab{}.
\newblock \showarticletitle{Chain of hindsight aligns language models with feedback}.
\newblock \bibinfo{journal}{\emph{arXiv preprint arXiv:2302.02676}} (\bibinfo{year}{2023}).
\newblock
\href{https://doi.org/10.48550/arXiv.2302.02676}{doi:\nolinkurl{10.48550/arXiv.2302.02676}}


\bibitem[Ma et~al\mbox{.}(2018)]%
        {ma2018mode}
\bibfield{author}{\bibinfo{person}{Shiqing Ma}, \bibinfo{person}{Yingqi Liu}, \bibinfo{person}{Wen-Chuan Lee}, \bibinfo{person}{Xiangyu Zhang}, {and} \bibinfo{person}{Ananth Grama}.} \bibinfo{year}{2018}\natexlab{}.
\newblock \showarticletitle{MODE: automated neural network model debugging via state differential analysis and input selection}. In \bibinfo{booktitle}{\emph{Proceedings of the 2018 26th ACM Joint Meeting on European Software Engineering Conference and Symposium on the Foundations of Software Engineering}}. \bibinfo{pages}{175--186}.
\newblock
\href{https://doi.org/10.1145/3236024.3236082}{doi:\nolinkurl{10.1145/3236024.3236082}}


\bibitem[Mechtaev et~al\mbox{.}(2015)]%
        {mechtaev2015directfix}
\bibfield{author}{\bibinfo{person}{Sergey Mechtaev}, \bibinfo{person}{Jooyong Yi}, {and} \bibinfo{person}{Abhik Roychoudhury}.} \bibinfo{year}{2015}\natexlab{}.
\newblock \showarticletitle{Directfix: Looking for simple program repairs}. In \bibinfo{booktitle}{\emph{2015 IEEE/ACM 37th IEEE International Conference on Software Engineering}}, Vol.~\bibinfo{volume}{1}. IEEE, \bibinfo{pages}{448--458}.
\newblock
\href{https://doi.org/10.1109/ICSE.2015.63}{doi:\nolinkurl{10.1109/ICSE.2015.63}}


\bibitem[Meng et~al\mbox{.}(2022)]%
        {meng2022locating}
\bibfield{author}{\bibinfo{person}{Kevin Meng}, \bibinfo{person}{David Bau}, \bibinfo{person}{Alex Andonian}, {and} \bibinfo{person}{Yonatan Belinkov}.} \bibinfo{year}{2022}\natexlab{}.
\newblock \showarticletitle{Locating and editing factual associations in GPT}.
\newblock \bibinfo{journal}{\emph{Advances in Neural Information Processing Systems}}  \bibinfo{volume}{35} (\bibinfo{year}{2022}), \bibinfo{pages}{17359--17372}.
\newblock


\bibitem[Merity et~al\mbox{.}(2017)]%
        {merity2016pointer}
\bibfield{author}{\bibinfo{person}{Stephen Merity}, \bibinfo{person}{Caiming Xiong}, \bibinfo{person}{James Bradbury}, {and} \bibinfo{person}{Richard Socher}.} \bibinfo{year}{2017}\natexlab{}.
\newblock \showarticletitle{Pointer Sentinel Mixture Models}. In \bibinfo{booktitle}{\emph{International Conference on Learning Representations}}.
\newblock
\urldef\tempurl%
\url{https://openreview.net/forum?id=Byj72udxe}
\showURL{%
\tempurl}


\bibitem[Mitchell et~al\mbox{.}(2021)]%
        {mitchell2021fast}
\bibfield{author}{\bibinfo{person}{Eric Mitchell}, \bibinfo{person}{Charles Lin}, \bibinfo{person}{Antoine Bosselut}, \bibinfo{person}{Chelsea Finn}, {and} \bibinfo{person}{Christopher~D Manning}.} \bibinfo{year}{2021}\natexlab{}.
\newblock \showarticletitle{Fast model editing at scale}.
\newblock \bibinfo{journal}{\emph{arXiv preprint arXiv:2110.11309}} (\bibinfo{year}{2021}).
\newblock
\href{https://doi.org/10.48550/arXiv.2110.11309}{doi:\nolinkurl{10.48550/arXiv.2110.11309}}


\bibitem[Nguyen et~al\mbox{.}(2013)]%
        {nguyen2013semfix}
\bibfield{author}{\bibinfo{person}{Hoang Duong~Thien Nguyen}, \bibinfo{person}{Dawei Qi}, \bibinfo{person}{Abhik Roychoudhury}, {and} \bibinfo{person}{Satish Chandra}.} \bibinfo{year}{2013}\natexlab{}.
\newblock \showarticletitle{Semfix: Program repair via semantic analysis}. In \bibinfo{booktitle}{\emph{2013 35th International Conference on Software Engineering (ICSE)}}. IEEE, \bibinfo{pages}{772--781}.
\newblock
\href{https://doi.org/10.1109/ICSE.2013.6606623}{doi:\nolinkurl{10.1109/ICSE.2013.6606623}}


\bibitem[Niu et~al\mbox{.}(2024)]%
        {niu2024parameter}
\bibfield{author}{\bibinfo{person}{Tong Niu}, \bibinfo{person}{Caiming Xiong}, \bibinfo{person}{Semih Yavuz}, {and} \bibinfo{person}{Yingbo Zhou}.} \bibinfo{year}{2024}\natexlab{}.
\newblock \showarticletitle{Parameter-Efficient Detoxification with Contrastive Decoding}.
\newblock \bibinfo{journal}{\emph{arXiv preprint arXiv:2401.06947}} (\bibinfo{year}{2024}).
\newblock
\href{https://doi.org/10.48550/arXiv.2401.06947}{doi:\nolinkurl{10.48550/arXiv.2401.06947}}


\bibitem[Ouyang et~al\mbox{.}(2022)]%
        {ouyang2022training}
\bibfield{author}{\bibinfo{person}{Long Ouyang}, \bibinfo{person}{Jeffrey Wu}, \bibinfo{person}{Xu Jiang}, \bibinfo{person}{Diogo Almeida}, \bibinfo{person}{Carroll Wainwright}, \bibinfo{person}{Pamela Mishkin}, \bibinfo{person}{Chong Zhang}, \bibinfo{person}{Sandhini Agarwal}, \bibinfo{person}{Katarina Slama}, \bibinfo{person}{Alex Ray}, {et~al\mbox{.}}} \bibinfo{year}{2022}\natexlab{}.
\newblock \showarticletitle{Training language models to follow instructions with human feedback}.
\newblock \bibinfo{journal}{\emph{Advances in neural information processing systems}}  \bibinfo{volume}{35} (\bibinfo{year}{2022}), \bibinfo{pages}{27730--27744}.
\newblock


\bibitem[Pan et~al\mbox{.}(2023)]%
        {pan2023automatically}
\bibfield{author}{\bibinfo{person}{Liangming Pan}, \bibinfo{person}{Michael Saxon}, \bibinfo{person}{Wenda Xu}, \bibinfo{person}{Deepak Nathani}, \bibinfo{person}{Xinyi Wang}, {and} \bibinfo{person}{William~Yang Wang}.} \bibinfo{year}{2023}\natexlab{}.
\newblock \showarticletitle{Automatically correcting large language models: Surveying the landscape of diverse self-correction strategies}.
\newblock \bibinfo{journal}{\emph{arXiv preprint arXiv:2308.03188}} (\bibinfo{year}{2023}).
\newblock
\href{https://doi.org/10.48550/arXiv.2308.03188}{doi:\nolinkurl{10.48550/arXiv.2308.03188}}


\bibitem[Paperno et~al\mbox{.}(2016)]%
        {paperno2016lambada}
\bibfield{author}{\bibinfo{person}{Denis Paperno}, \bibinfo{person}{Germ{\'a}n Kruszewski}, \bibinfo{person}{Angeliki Lazaridou}, \bibinfo{person}{Ngoc-Quan Pham}, \bibinfo{person}{Raffaella Bernardi}, \bibinfo{person}{Sandro Pezzelle}, \bibinfo{person}{Marco Baroni}, \bibinfo{person}{Gemma Boleda}, {and} \bibinfo{person}{Raquel Fern{\'a}ndez}.} \bibinfo{year}{2016}\natexlab{}.
\newblock \showarticletitle{The LAMBADA dataset: Word prediction requiring a broad discourse context}. In \bibinfo{booktitle}{\emph{Proceedings of the 54th Annual Meeting of the Association for Computational Linguistics (Volume 1: Long Papers)}}. \bibinfo{pages}{1525--1534}.
\newblock
\href{https://doi.org/10.48550/arXiv.1606.06031}{doi:\nolinkurl{10.48550/arXiv.1606.06031}}


\bibitem[Paszke et~al\mbox{.}(2017)]%
        {paszke2017automatic}
\bibfield{author}{\bibinfo{person}{Adam Paszke}, \bibinfo{person}{Sam Gross}, \bibinfo{person}{Soumith Chintala}, \bibinfo{person}{Gregory Chanan}, \bibinfo{person}{Edward Yang}, \bibinfo{person}{Zachary DeVito}, \bibinfo{person}{Zeming Lin}, \bibinfo{person}{Alban Desmaison}, \bibinfo{person}{Luca Antiga}, {and} \bibinfo{person}{Adam Lerer}.} \bibinfo{year}{2017}\natexlab{}.
\newblock \showarticletitle{Automatic differentiation in pytorch}.
\newblock  (\bibinfo{year}{2017}).
\newblock


\bibitem[Radford et~al\mbox{.}(2019)]%
        {radford2019language}
\bibfield{author}{\bibinfo{person}{Alec Radford}, \bibinfo{person}{Jeffrey Wu}, \bibinfo{person}{Rewon Child}, \bibinfo{person}{David Luan}, \bibinfo{person}{Dario Amodei}, \bibinfo{person}{Ilya Sutskever}, {et~al\mbox{.}}} \bibinfo{year}{2019}\natexlab{}.
\newblock \showarticletitle{Language models are unsupervised multitask learners}.
\newblock \bibinfo{journal}{\emph{OpenAI blog}} \bibinfo{volume}{1}, \bibinfo{number}{8} (\bibinfo{year}{2019}), \bibinfo{pages}{9}.
\newblock


\bibitem[Rafailov et~al\mbox{.}(2024)]%
        {rafailov2024direct}
\bibfield{author}{\bibinfo{person}{Rafael Rafailov}, \bibinfo{person}{Archit Sharma}, \bibinfo{person}{Eric Mitchell}, \bibinfo{person}{Christopher~D Manning}, \bibinfo{person}{Stefano Ermon}, {and} \bibinfo{person}{Chelsea Finn}.} \bibinfo{year}{2024}\natexlab{}.
\newblock \showarticletitle{Direct preference optimization: Your language model is secretly a reward model}.
\newblock \bibinfo{journal}{\emph{Advances in Neural Information Processing Systems}}  \bibinfo{volume}{36} (\bibinfo{year}{2024}).
\newblock


\bibitem[Schick et~al\mbox{.}(2021)]%
        {schick2021self}
\bibfield{author}{\bibinfo{person}{Timo Schick}, \bibinfo{person}{Sahana Udupa}, {and} \bibinfo{person}{Hinrich Sch{\"u}tze}.} \bibinfo{year}{2021}\natexlab{}.
\newblock \showarticletitle{Self-diagnosis and self-debiasing: A proposal for reducing corpus-based bias in nlp}.
\newblock \bibinfo{journal}{\emph{Transactions of the Association for Computational Linguistics}}  \bibinfo{volume}{9} (\bibinfo{year}{2021}), \bibinfo{pages}{1408--1424}.
\newblock
\href{https://doi.org/10.1162/tacl_a_00434}{doi:\nolinkurl{10.1162/tacl_a_00434}}


\bibitem[Shoeybi et~al\mbox{.}(2019)]%
        {shoeybi2019megatron}
\bibfield{author}{\bibinfo{person}{Mohammad Shoeybi}, \bibinfo{person}{Mostofa Patwary}, \bibinfo{person}{Raul Puri}, \bibinfo{person}{Patrick LeGresley}, \bibinfo{person}{Jared Casper}, {and} \bibinfo{person}{Bryan Catanzaro}.} \bibinfo{year}{2019}\natexlab{}.
\newblock \showarticletitle{Megatron-lm: Training multi-billion parameter language models using model parallelism}.
\newblock \bibinfo{journal}{\emph{arXiv preprint arXiv:1909.08053}} (\bibinfo{year}{2019}).
\newblock
\href{https://doi.org/10.48550/arXiv.1909.08053}{doi:\nolinkurl{10.48550/arXiv.1909.08053}}


\bibitem[Si et~al\mbox{.}(2022)]%
        {si2022prompting}
\bibfield{author}{\bibinfo{person}{Chenglei Si}, \bibinfo{person}{Zhe Gan}, \bibinfo{person}{Zhengyuan Yang}, \bibinfo{person}{Shuohang Wang}, \bibinfo{person}{Jianfeng Wang}, \bibinfo{person}{Jordan Boyd-Graber}, {and} \bibinfo{person}{Lijuan Wang}.} \bibinfo{year}{2022}\natexlab{}.
\newblock \showarticletitle{Prompting gpt-3 to be reliable}.
\newblock \bibinfo{journal}{\emph{arXiv preprint arXiv:2210.09150}} (\bibinfo{year}{2022}).
\newblock
\href{https://doi.org/10.48550/arXiv.2210.09150}{doi:\nolinkurl{10.48550/arXiv.2210.09150}}


\bibitem[Solaiman and Dennison(2021)]%
        {solaiman2021process}
\bibfield{author}{\bibinfo{person}{Irene Solaiman} {and} \bibinfo{person}{Christy Dennison}.} \bibinfo{year}{2021}\natexlab{}.
\newblock \showarticletitle{Process for adapting language models to society (palms) with values-targeted datasets}.
\newblock \bibinfo{journal}{\emph{Advances in Neural Information Processing Systems}}  \bibinfo{volume}{34} (\bibinfo{year}{2021}), \bibinfo{pages}{5861--5873}.
\newblock


\bibitem[Vaswani et~al\mbox{.}(2017)]%
        {vaswani2017attention}
\bibfield{author}{\bibinfo{person}{Ashish Vaswani}, \bibinfo{person}{Noam Shazeer}, \bibinfo{person}{Niki Parmar}, \bibinfo{person}{Jakob Uszkoreit}, \bibinfo{person}{Llion Jones}, \bibinfo{person}{Aidan~N Gomez}, \bibinfo{person}{{\L}ukasz Kaiser}, {and} \bibinfo{person}{Illia Polosukhin}.} \bibinfo{year}{2017}\natexlab{}.
\newblock \showarticletitle{Attention is all you need}.
\newblock \bibinfo{journal}{\emph{Advances in neural information processing systems}}  \bibinfo{volume}{30} (\bibinfo{year}{2017}).
\newblock


\bibitem[Wang et~al\mbox{.}(2022)]%
        {wang2022exploring}
\bibfield{author}{\bibinfo{person}{Boxin Wang}, \bibinfo{person}{Wei Ping}, \bibinfo{person}{Chaowei Xiao}, \bibinfo{person}{Peng Xu}, \bibinfo{person}{Mostofa Patwary}, \bibinfo{person}{Mohammad Shoeybi}, \bibinfo{person}{Bo Li}, \bibinfo{person}{Anima Anandkumar}, {and} \bibinfo{person}{Bryan Catanzaro}.} \bibinfo{year}{2022}\natexlab{}.
\newblock \showarticletitle{Exploring the limits of domain-adaptive training for detoxifying large-scale language models}.
\newblock \bibinfo{journal}{\emph{Advances in Neural Information Processing Systems}}  \bibinfo{volume}{35} (\bibinfo{year}{2022}), \bibinfo{pages}{35811--35824}.
\newblock


\bibitem[Wang et~al\mbox{.}(2024)]%
        {wang2024detoxifying}
\bibfield{author}{\bibinfo{person}{Mengru Wang}, \bibinfo{person}{Ningyu Zhang}, \bibinfo{person}{Ziwen Xu}, \bibinfo{person}{Zekun Xi}, \bibinfo{person}{Shumin Deng}, \bibinfo{person}{Yunzhi Yao}, \bibinfo{person}{Qishen Zhang}, \bibinfo{person}{Linyi Yang}, \bibinfo{person}{Jindong Wang}, {and} \bibinfo{person}{Huajun Chen}.} \bibinfo{year}{2024}\natexlab{}.
\newblock \showarticletitle{Detoxifying Large Language Models via Knowledge Editing}.
\newblock \bibinfo{journal}{\emph{arXiv preprint arXiv:2403.14472}} (\bibinfo{year}{2024}).
\newblock
\href{https://doi.org/10.48550/arXiv.2403.14472}{doi:\nolinkurl{10.48550/arXiv.2403.14472}}


\bibitem[Wardat et~al\mbox{.}(2022)]%
        {wardat2022deepdiagnosis}
\bibfield{author}{\bibinfo{person}{Mohammad Wardat}, \bibinfo{person}{Breno~Dantas Cruz}, \bibinfo{person}{Wei Le}, {and} \bibinfo{person}{Hridesh Rajan}.} \bibinfo{year}{2022}\natexlab{}.
\newblock \showarticletitle{Deepdiagnosis: automatically diagnosing faults and recommending actionable fixes in deep learning programs}. In \bibinfo{booktitle}{\emph{Proceedings of the 44th international conference on software engineering}}. \bibinfo{pages}{561--572}.
\newblock
\href{https://doi.org/10.1145/3510003.3510071}{doi:\nolinkurl{10.1145/3510003.3510071}}


\bibitem[Weiser(1984)]%
        {weiser1984program}
\bibfield{author}{\bibinfo{person}{Mark Weiser}.} \bibinfo{year}{1984}\natexlab{}.
\newblock \showarticletitle{Program slicing}.
\newblock \bibinfo{journal}{\emph{IEEE Transactions on software engineering}} \bibinfo{number}{4} (\bibinfo{year}{1984}), \bibinfo{pages}{352--357}.
\newblock


\bibitem[Wen et~al\mbox{.}(2018)]%
        {wen2018context}
\bibfield{author}{\bibinfo{person}{Ming Wen}, \bibinfo{person}{Jun~jie Chen}, \bibinfo{person}{Rongxin Wu}, \bibinfo{person}{Dan Hao}, {and} \bibinfo{person}{Shing-Chi Cheung}.} \bibinfo{year}{2018}\natexlab{}.
\newblock \showarticletitle{Context-aware patch generation for better automated program repair}. In \bibinfo{booktitle}{\emph{Proceedings of the 40th international conference on software engineering}}. \bibinfo{pages}{1--11}.
\newblock
\href{https://doi.org/10.1145/3180155.318023}{doi:\nolinkurl{10.1145/3180155.318023}}


\bibitem[Weng et~al\mbox{.}(2023)]%
        {weng2023large}
\bibfield{author}{\bibinfo{person}{Yixuan Weng}, \bibinfo{person}{Minjun Zhu}, \bibinfo{person}{Fei Xia}, \bibinfo{person}{Bin Li}, \bibinfo{person}{Shizhu He}, \bibinfo{person}{Shengping Liu}, \bibinfo{person}{Bin Sun}, \bibinfo{person}{Kang Liu}, {and} \bibinfo{person}{Jun Zhao}.} \bibinfo{year}{2023}\natexlab{}.
\newblock \showarticletitle{Large language models are better reasoners with self-verification}. In \bibinfo{booktitle}{\emph{Findings of the Association for Computational Linguistics: EMNLP 2023}}. \bibinfo{pages}{2550--2575}.
\newblock
\href{https://doi.org/10.18653/v1/2023.findings-emnlp.167}{doi:\nolinkurl{10.18653/v1/2023.findings-emnlp.167}}


\bibitem[Xie et~al\mbox{.}(2023)]%
        {xie2023defending}
\bibfield{author}{\bibinfo{person}{Yueqi Xie}, \bibinfo{person}{Jingwei Yi}, \bibinfo{person}{Jiawei Shao}, \bibinfo{person}{Justin Curl}, \bibinfo{person}{Lingjuan Lyu}, \bibinfo{person}{Qifeng Chen}, \bibinfo{person}{Xing Xie}, {and} \bibinfo{person}{Fangzhao Wu}.} \bibinfo{year}{2023}\natexlab{}.
\newblock \showarticletitle{Defending chatgpt against jailbreak attack via self-reminders}.
\newblock \bibinfo{journal}{\emph{Nature Machine Intelligence}} \bibinfo{volume}{5}, \bibinfo{number}{12} (\bibinfo{year}{2023}), \bibinfo{pages}{1486--1496}.
\newblock
\href{https://doi.org/10.1038/s42256-023-00765-8}{doi:\nolinkurl{10.1038/s42256-023-00765-8}}


\bibitem[Xu et~al\mbox{.}(2022)]%
        {xu2022leashing}
\bibfield{author}{\bibinfo{person}{Canwen Xu}, \bibinfo{person}{Zexue He}, \bibinfo{person}{Zhankui He}, {and} \bibinfo{person}{Julian McAuley}.} \bibinfo{year}{2022}\natexlab{}.
\newblock \showarticletitle{Leashing the inner demons: Self-detoxification for language models}. In \bibinfo{booktitle}{\emph{Proceedings of the AAAI Conference on Artificial Intelligence}}, Vol.~\bibinfo{volume}{36}. \bibinfo{pages}{11530--11537}.
\newblock
\href{https://doi.org/10.1609/aaai.v36i10.21406}{doi:\nolinkurl{10.1609/aaai.v36i10.21406}}


\bibitem[Yang et~al\mbox{.}(2022)]%
        {yang2022unified}
\bibfield{author}{\bibinfo{person}{Zonghan Yang}, \bibinfo{person}{Xiaoyuan Yi}, \bibinfo{person}{Peng Li}, \bibinfo{person}{Yang Liu}, {and} \bibinfo{person}{Xing Xie}.} \bibinfo{year}{2022}\natexlab{}.
\newblock \showarticletitle{Unified detoxifying and debiasing in language generation via inference-time adaptive optimization}.
\newblock \bibinfo{journal}{\emph{arXiv preprint arXiv:2210.04492}} (\bibinfo{year}{2022}).
\newblock
\href{https://doi.org/10.48550/arXiv.2210.04492}{doi:\nolinkurl{10.48550/arXiv.2210.04492}}


\bibitem[Zhang and Chan(2019)]%
        {zhang2019apricot}
\bibfield{author}{\bibinfo{person}{Hao Zhang} {and} \bibinfo{person}{WK Chan}.} \bibinfo{year}{2019}\natexlab{}.
\newblock \showarticletitle{Apricot: A weight-adaptation approach to fixing deep learning models}. In \bibinfo{booktitle}{\emph{2019 34th IEEE/ACM International Conference on Automated Software Engineering (ASE)}}. IEEE, \bibinfo{pages}{376--387}.
\newblock
\href{https://doi.org/10.1109/ASE.2019.00043}{doi:\nolinkurl{10.1109/ASE.2019.00043}}


\bibitem[Zhang et~al\mbox{.}(2021)]%
        {zhang2021autotrainer}
\bibfield{author}{\bibinfo{person}{Xiaoyu Zhang}, \bibinfo{person}{Juan Zhai}, \bibinfo{person}{Shiqing Ma}, {and} \bibinfo{person}{Chao Shen}.} \bibinfo{year}{2021}\natexlab{}.
\newblock \showarticletitle{Autotrainer: An automatic dnn training problem detection and repair system}. In \bibinfo{booktitle}{\emph{2021 IEEE/ACM 43rd International Conference on Software Engineering (ICSE)}}. IEEE, \bibinfo{pages}{359--371}.
\newblock
\href{https://doi.org/10.1109/ICSE43902.2021.00043}{doi:\nolinkurl{10.1109/ICSE43902.2021.00043}}


\bibitem[Zhang et~al\mbox{.}(2020)]%
        {zhang2020dynamic}
\bibfield{author}{\bibinfo{person}{Ziqi Zhang}, \bibinfo{person}{Yuanchun Li}, \bibinfo{person}{Yao Guo}, \bibinfo{person}{Xiangqun Chen}, {and} \bibinfo{person}{Yunxin Liu}.} \bibinfo{year}{2020}\natexlab{}.
\newblock \showarticletitle{Dynamic slicing for deep neural networks}. In \bibinfo{booktitle}{\emph{Proceedings of the 28th ACM Joint Meeting on European Software Engineering Conference and Symposium on the Foundations of Software Engineering}}. \bibinfo{pages}{838--850}.
\newblock
\href{https://doi.org/10.1145/3368089.3409676}{doi:\nolinkurl{10.1145/3368089.3409676}}


\bibitem[Zhang et~al\mbox{.}(2022)]%
        {zhang2022remos}
\bibfield{author}{\bibinfo{person}{Ziqi Zhang}, \bibinfo{person}{Yuanchun Li}, \bibinfo{person}{Jindong Wang}, \bibinfo{person}{Bingyan Liu}, \bibinfo{person}{Ding Li}, \bibinfo{person}{Yao Guo}, \bibinfo{person}{Xiangqun Chen}, {and} \bibinfo{person}{Yunxin Liu}.} \bibinfo{year}{2022}\natexlab{}.
\newblock \showarticletitle{ReMoS: reducing defect inheritance in transfer learning via relevant model slicing}. In \bibinfo{booktitle}{\emph{Proceedings of the 44th International Conference on Software Engineering}}. \bibinfo{pages}{1856--1868}.
\newblock
\href{https://doi.org/10.1145/3510003.3510191}{doi:\nolinkurl{10.1145/3510003.3510191}}


\bibitem[Zhao et~al\mbox{.}(2023)]%
        {zhao2023survey}
\bibfield{author}{\bibinfo{person}{Wayne~Xin Zhao}, \bibinfo{person}{Kun Zhou}, \bibinfo{person}{Junyi Li}, \bibinfo{person}{Tianyi Tang}, \bibinfo{person}{Xiaolei Wang}, \bibinfo{person}{Yupeng Hou}, \bibinfo{person}{Yingqian Min}, \bibinfo{person}{Beichen Zhang}, \bibinfo{person}{Junjie Zhang}, \bibinfo{person}{Zican Dong}, {et~al\mbox{.}}} \bibinfo{year}{2023}\natexlab{}.
\newblock \showarticletitle{A survey of large language models}.
\newblock \bibinfo{journal}{\emph{arXiv preprint arXiv:2303.18223}} (\bibinfo{year}{2023}).
\newblock
\href{https://doi.org/10.48550/arXiv.2303.18223}{doi:\nolinkurl{10.48550/arXiv.2303.18223}}


\end{thebibliography}


\end{document}